\documentclass[preprint,authoryear,review,11pt]{elsarticle}

\usepackage{graphicx}

\usepackage{amssymb}

\usepackage[top=4cm,left=3cm,right=3cm,bottom=4cm]{geometry}

\usepackage{enumitem} 
  \setlist{itemsep=1ex plus0.2ex, leftmargin=*, align=left}

\makeatletter
\newcommand{\labitem}[2]{%
\def\@itemlabel{\textbf{#1}}
\item
\def\@currentlabel{#1}\label{#2}}
\makeatother

\makeatletter
\newcommand{\headingitem}[1]{%
\vspace{0.3cm}
\def\@itemlabel{\textbf{#1}}
\item
\def\@currentlabel{#1}
\addtocounter{enumi}{-1}
}
\makeatother

\usepackage{rotating}
\usepackage{csvsimple}
\usepackage{multirow}
\usepackage{lipsum}
\usepackage{afterpage}

\usepackage{etoolbox}
\robustify\bfseries

\usepackage{eurosym}
\usepackage{siunitx}

    \DeclareSIUnit\eur{\officialeuro}
    \DeclareSIUnit\M{M}
    \DeclareSIUnit\k{k}

  \def\sym#1{\ifmmode^{#1}\else\(^{#1}\)\fi}

\usepackage{setspace}
\usepackage{csquotes}
\usepackage{amsmath}
\usepackage{bm}
  
\usepackage{mathtools} 
\usepackage{bbm}

\usepackage{xspace}
	\newcommand\ie{i.\,e.\xspace}
	\newcommand\eg{e.\,g.\xspace}

	\newcommand\cf{cf.\xspace}

	\newcommand\iid{i.\,i.\,d.\xspace}

\usepackage[amsmath,hyperref,framed]{ntheorem} 
  \theoremstyle{plain}
     \newtheorem{theorem}{Theorem}[section]
    \newtheorem{remark}[theorem]{Remark}
     \newtheorem{definition}[theorem]{Definition}
     
  \theoremstyle{nonumberplain}
    \theoremseparator{.}
    \theoremheaderfont{\bfseries}
    \theorembodyfont{\normalfont}
    \theoremsymbol{$\blacksquare$}
      \RequirePackage{amssymb}

      \makeatletter
    \let\copy@theorem@headerfont=\theorem@headerfont
    \newcommand{\my@theorem@headerfont}{%
        \boldmath\copy@theorem@headerfont\unboldmath
      }
    \let\theorem@headerfont=\my@theorem@headerfont
      \makeatother

\theoremstyle{nonumberplain}
\theoremseparator{.}
\newcommand{\argmax}{\operatornamewithlimits{arg \, max}}
\newcommand{\argmin}{\operatornamewithlimits{arg \, min}}

\usepackage[%
  sort&compress
]{cleveref}

\usepackage{url}

\usepackage{isomath}

\newcommand\transpose{^T} 

  \newcommand\define{\coloneqq} 

  \newcommand{\norm}[1]{\left\lVert#1\right\rVert}
 
\newcommand{\domain}{\mathcal{D}}
\newcommand{\R}{\mathbb{R}}

\newcommand{\alg}{\mathcal{A}}
\newcommand{\indicator}{\mathbbm{1}}
\newcommand{\expectation}{\mathbb{E}}
\newcommand{\normal}{\mathcal{N}}
\newcommand{\unif}{\mathcal{U}}
\newcommand{\GP}{\mathcal{GP}}
\newcommand{\Python}{\texttt{Python}}
\newcommand{\npart}{n_\text{par}}

\newcommand{\memory}[1]{\mathcal{M}^{(#1)}}

\newcommand{\particles}[1]{\mathcal{W}^{(#1)}}
\newcommand{\position}[2]{x^{(#2)}_#1}
\newcommand{\velocity}[2]{v^{(#2)}_#1}
\newcommand{\personal}[2]{p^{(#2)}_#1}

\newcommand{\intervalcc}[2]{\left[#1,#2\right]}

\usepackage{algorithm}
\usepackage{algorithmic}        

    \Crefname{ALC@unique}{step}{steps}
    \Crefname{ALC@unique}{Step}{Steps}
    \Crefname{ALC@line}{step}{steps}
    \Crefname{ALC@line}{Step}{Steps}
    
\usepackage{tabularx}
\usepackage{ltablex}
\usepackage[usenames,dvipsnames]{xcolor}

\usepackage{colortbl}
\usepackage{tabularx}
\usepackage{booktabs}
\usepackage{collcell}

\usepackage{placeins}

\usepackage{ragged2e}
  
\newcommand{\PreserveBackslash}[1]{\let\temp=\\#1\let\\=\temp}
\newcolumntype{v}[1]{>{\PreserveBackslash\RaggedRight\hspace{0pt}}p{#1}}

  \usepackage{adjustbox}
\newcolumntype{Q}[2]{%
    >{\adjustbox{angle=#1,lap=\width-(#2)}\bgroup}%
    l%
    <{\egroup}%
}

\newcommand{\mcellt}[2][c]{%
  \begin{tabular}[t]{@{}#1@{}}#2\end{tabular}}

\usepackage[
    colorinlistoftodos,
    textsize=footnotesize,
        ]{todonotes}

\makeatletter
    \renewcommand{\fps@figure}{htb}         
    \renewcommand{\fps@table}{htb}         
\makeatother 

\usepackage{float}
\usepackage{rotating}

\usepackage{longtable}
\usepackage{tabu}
\usepackage{multibib}
\usepackage{floatpag}

\usepackage{multibib}
\newcites{sec}{References for Online Appendix}

\journal{arXiv}

\begin{document}

\begin{frontmatter}



\title{Directed particle swarm optimization with Gaussian-process-based function forecasting}


\author[ETH,KIT]{Johannes Jakubik}
\ead{johannes.jakubik@kit.edu}

\author[ETH]{Adrian Binding}

\author[ETH]{Stefan Feuerriegel\corref{cor1}}
\ead{sfeuerriegel@ethz.ch}

\address[ETH]{ETH Zurich, Weinbergstr. 56/58, 8092 Zurich, Switzerland}
\address[KIT]{Karlsruhe Institute of Technology, Kaiserstraße 12,  76131 Karlsruhe, Germany}

\cortext[cor1]{Corresponding author.}

\begin{abstract}
Particle swarm optimization (PSO) is an iterative search method that moves a set of candidate solution around a search-space towards the best known global and local solutions with randomized step lengths. PSO frequently accelerates optimization in practical applications, where gradients are not available and function evaluations expensive. Yet the traditional PSO algorithm ignores the potential knowledge that could have been gained of the objective function from the observations by individual particles. Hence, we draw upon concepts from Bayesian optimization and introduce a stochastic surrogate model of the objective function. That is, we fit a Gaussian process to past evaluations of the objective function, forecast its shape and then adapt the particle movements based on it. Our computational experiments demonstrate that baseline implementations of PSO (\ie, SPSO2011) are outperformed. Furthermore, compared to, state-of-art surrogate-assisted evolutionary algorithms, we achieve substantial performance improvements on several popular benchmark functions. Overall, we find that our algorithm attains desirable properties for exploratory and exploitative behavior.
\end{abstract}

\end{frontmatter}

\sloppy
\raggedbottom

\newpage
\section{Introduction}


Stochastic optimization methods refer to optimization methods that incorporate random variables into a search process \citep[Chapter~7]{Gentle.2012} and often improves the performance in a large variety of practical settings \citep{Hoos.2005}. Stochastic optimization methods are frequently utilized in black-box optimization settings, in particular for derivative-free optimization \citep{Pham.2014,Rios.2013}. Here no assumptions regarding the analytic form of the objective function are made and, because of it, gradients are unavailable. That is, one can only query a function $f$ for single points $x$, for which the corresponding evaluation $f(x)$ is then returned. Such problems are prevalent in numerous applications from engineering, medicine and economics among others, where the underlying function is computationally or economically expensive to evaluate \citep{Rios.2013}.


A prominent example of stochastic search methods is particle swarm optimization~(PSO), which presents the focus of this paper. The idea behind PSO stems from population-based simulations of animal behavior \citep{Kennedy.1995}. As such, a set of candidate solutions (\ie, particles) are initialized at random positions, which are then moved around a search-space. These then proceed towards a mixture of the best-known swarm position and the particle's best location from its past trajectory. Several variants to the original PSO algorithm have been developed \citep[\eg, SPSO2011;][]{ZambranoBigiarini.2013}, which we summarize in our review section. For detailed surveys of PSO variants, we refer to \citet{Poli.2007} and \citet{Bonyadi.2017}.


PSO is straightforward to implement and makes little assumptions on the underlying optimization problem. In particular, the optimization is derivative-free and can also adapt to non-convex problems. As a result, it is used for multi-objective optimization \citep[\eg,][]{Liu.2017,zouache2018cooperative,yu2017multi,Sethanan.2016}, as well as in various areas of application, such as energy \citep{yu2017multi}, supply chain management \citep{hong2018optimal}, and operations research in general \citep{Etgar.2017, Tasgetiren.2007}. PSO can further be extended to unconstrained optimization problems \citep{Bonyadi.2017}, as well as integer programming \citep{Laskari.2002}. Convergence guarantees can be made for a wide range of settings \citep{Jiang.2007}. 


The information actually utilized by PSO and its variants is limited: each particle is merely controlled by its personal best position, as well as the best of a selection of other particles. In other words, no knowledge regarding the past trajectories in the search-space are retained and all corresponding function evaluations are \textquote{lost}. As a result, particles might re-visit a neighborhood that has previously been explored; thereby leading to potentially redundant function calls \citep{Iqbal.2006}. On the other hand, all particle movements are guided towards the best \textit{known} positions. This strategy could miss a global optimum that is hidden in an unexplored region \citep[cf.][]{Pham.2014} and may lead to convergence to a non-optimum \citep{vandenBergh.2002}.


This work aims at improving the search strategy that controls the movements of the particle swarm. We thus propose a combination of the PSO methodology, where the swarm intelligence leverages a stochastic surrogate model of the objective function. This allows us to estimate the surface of the objective function (\ie the landscape) from the past search trajectory of the particles. Based on this surrogate model, we can then enhance particle movements in the search process with respect to exploratory and exploitative behavior. More specifically, we develop a variant (A)~in which we modify the movements for all particles by incorporating a third direction pointing to the global optima of the surrogate. Further variants re-locate individual particles in order to let them (B)~exploit directly global optima suggested by the surrogate model and (C)~explore regions with high uncertainty in the surrogate. 


Mathematically, we draw upon Gaussian processes (GPs) as stochastic model for the surrogate. This choice is motivated by common procedures in Bayesian optimization \citep{Snoek.2012} and response surface methods via kriging \citep{Rios.2013}. In addition, it provides rigorous uncertainty estimates for the function approximation, thereby facilitating explorations of locations that promise the highest probability of improvement in the surrogate model. Our function approximation thus reveals similarities to Bayesian optimization, where one also formulates a stochastic model over the function space \citep{Mockus.1975}. However, one then chooses a sequence of points in order to improve the overall objective, as if a single particle jumps through the search space (see \ref{appendix:bayesian_optimization} for details). Different from PSO, these movements are deterministic and can thus not benefit from a randomization in the optimization method.


We finally perform a series of computational experiments in order to demonstrate the improved convergence of our proposed algorithm. For this purpose, we adhere to earlier works that develop modifications to particle swarm optimization, that is, we use the SPSO2011 implementation \citep{ZambranoBigiarini.2013}. We further follow the literature with regard to the evaluation setting, that is, we use the CEC2013 suite of benchmark functions \citep{liang2013problem, ZambranoBigiarini.2013} that is common for evaluating PSO approaches. Our algorithm can outperform SPSO2011 for the entire set of benchmark functions. We find considerable improvements in the convergence in early stages of the iterative search. Statistical tests further demonstrate that the improvements are significant at common significance levels. 


Our work entails multiple contributions to the field of stochastic optimization. That is, we present an innovative combination of swarm search and machine learning tools (\ie, Gaussian process). Thereby, we contribute to the state-of-the-art by suggesting a statistical procedure to better leverage search trajectories in PSO. This thus introduces strategic control into the otherwise randomized search process. The resulting performance improvements stem from a better trade-off with respect to exploration and exploitation in the stochastic swarm search. This may prove useful in a wide variety of real-world optimization settings, where one utilizes PSO.  


The rest of this paper is structured as follows. \Cref{sec:PSO} reviews the mathematical specification of particle swarm optimization. Based on it, \Cref{sec:method} proposes our algorithm which combines PSO with a Gaussian-process-based surrogate model. \Cref{sec:results} then performs a series of computational experiments that demonstrate the acceleration in convergence of our proposed algorithms. \Cref{sec:discussion} discusses our algorithm, while \Cref{sec:conclusion} finally concludes.

\section{Background}
\label{sec:PSO}

Mathematically, we define black-box optimization problem as follows. Let $f: \domain \to \R$ denote an arbitrary (and potentially non-convex) function with a known (closed) domain $\domain$. Then our goal is to find $x^\ast \in \argmin_{x \in \domain} f(x)$, where we assume that $\domain \subseteq \R^p$ is connected and $f$ is continuous. In many applications, the underlying function is expensive to evaluate and, in these case, we might prefer to terminate the search process after a certain number of iterations or when the relative convergence fulfills predefined criteria. The last iteration of the algorithm can then be assessed via the expectation of a fixed loss function, for example, the mean squared loss
\begin{equation}
\expectation\left[  \norm{f(\alg(f)) - \min_{x \in \domain} f(x)}^2 \right] .
\label{eqn:expectedRegret}
\end{equation}

\subsection{Particle swarm optimization}


Black-box optimization is often approached by particle swarm optimization. In the following, we provide a brief specification of particle swarm optimization \citep[cf.][for a detailed overview]{Bonyadi.2017}. PSO stems from the simulation of socially-coordinated behavior of animal swarms \citep{Kennedy.1995}. It relies on a swarm of particles $j = 1, \ldots, \npart$ serving as candidate solutions. These particles move through the domain of the function in a coordinated fashion, where each particle is guided by individual experience, as well as observations of other particles. For this reason, each particle comes with information $w_j^{(i)} \define \left( x_j^{(i)}, v_j^{(i)}, p_j^{(i)} \right)$ in iteration $i$, where
\begin{quote}\begin{itemize}
\item $x_j^{(i)}$ is the position of particle $j$,
\item $v_j^{(i)}$ is its velocity, and
\item $p_j^{(i)}$ is its personal best position up to iteration $i$.
\end{itemize}\end{quote}
During the initialization phase, the particle positions $x_j^{(0)}$ are distributed across $\domain$. This initialization may be uniformly at random, or according to some other initialization scheme (\eg a Sobol sequence). The initial velocities $v_j^{(0)}$ are also chosen at random according to a predefined initialization procedure. 


In subsequent iterations, the particles move around the domain $\domain$, updating their respective velocity based on the personal best position $p_{j}^{(i)}$ and on the swarm intelligence. In the original PSO, the latter component is set via the global best-known position $g^{(i)} = \arg\min_{p_{j}^{(i)}}\;f(p_{j}^{(i)})$, while some PSO variants only consider a certain neighborhood $\mathcal{N}$ of particles, depending on the particle. This yields update rules
\begin{align}
v_j^{(i+1)} &= \mathit{update}_N \left(x_j^{(i)}, v_j^{(i)}, p_j^{(i)} \right) ,\\
x_j^{(i+1)} &= x_j^{(i)} + v_j^{(i+1)} , \\
p_j^{(i+1)} &= 
\begin{cases}
x_j^{(i)}, & \text{if } f( x_j^{(i+1)} ) < f( p_j^{(i)} ), \\
p_j^{(i)}, & \text{otherwise}.
\end{cases}
\end{align} 
Accordingly, the velocity update is key in controlling the movements of the swarm in space. In the original PSO~(OPSO), the velocity update is given by
\begin{equation}
\mathit{update}_N \left( v_j^{(i)}, p_j^{(i)} \right) = v_j^{(i)} + \phi_p \, R_{p,j}^{(i)}\odot (p_j^{(i)} - v_j^{(i)})
+ \phi_g \, R_{g,j}^{(i)} \odot (g^{(i)} - v_j^{(i)} )
\label{eqn:OPSO},
\end{equation}
where $R_{p,j}$ and $R_{g,j}$, are random vectors with components drawn from a uniform distribution $\unif(0, 1)$. This yields a linear combination for computing the new velocity with pre-defined coefficients $\phi_p$ and $\phi_g$ that additionally determine the relative influence of personal and global best directions and where $\odot$ denotes component-wise multiplication. The vectors $p_j^{(i)} - v_j^{(i)}$ and $g^{(i)} - v_j^{(i)}$ are referred to as the \emph{cognitive influence}~(CI) and \emph{social influence}~(SI), respectively.

\subsection{Extension to SPSO2011}



Various modifications to the above update rules have been proposed in the literature. For instance, PSO has been extended by adjustments to the method itself \citep[e.\,g.][]{Yin.2010} or combinations with other optimization schemes \citep[e.\,g.][]{Fan.2007}. For the sake of establishing a benchmark moving forward and focusing on reproducible insights, \citet{Bonyadi.2017} refer to the following variant as the standard PSO~(SPSO in short). Another standard version of PSO was introduced by \citet{ZambranoBigiarini.2013}, which is called SPSO2011. Instead of mixing component-wise uniformly random shifts in the velocity vector, particles are accelerated according to a hyper-spherical distribution. The velocity update is here given by
\begin{equation}
\mathit{update}_N \left( v_j^{(i)}, p_j^{(i)} \right) = \omega \, v_j^{(i)} + \mathcal{H}\left(G_j^{(i)}, \norm{G_j^{(i)} - x_j^{(i)}}\right) ,
\label{eqn:SPSO2011}
\end{equation}
where $\mathcal{H}\left(x, r\right)$ refers to as a hyperspherical distribution over the sphere with center $x$ and radius $r$, and $G_j^{(i)}$ is given as $G_j^{(i)} := x_j^{(i)} + \frac{\phi_p \text{CI} + \phi_g \text{SI}}{3}$ \citep{Bonyadi.2017}. SPSO2011 is regarded as state-of-the-art and should be used for benchmarking \citep{ZambranoBigiarini.2013}. Hence, it is later used in our experiments.

The update rules described above lead to convergence of the individual particles. This convergence behavior has been studied extensively \citep{Bonyadi.2017}. There is however no guarantee that the particles converge towards a global optimum.
Other variants exist, such as modifications  that replace the linear position update by arbitrary functions $x^{(i+1)}_j = \xi(x_j^{(i)}, v_j^{(i)})$ \citep[\cf\eg][]{vandenBergh.2002}, or which limit the neighborhood $N$ in specific ways \citep[\eg][]{Bratton.2007}. For a detailed review, we refer to \citet{Bonyadi.2017}. 

\subsection{Surrogate-assisted PSO}

PSO can often require numerous function calls until convergence, which might not be feasible for problems with expensive function evaluations. Here previous research has proposed to replace certain evaluations of the actual function $f$ by an approximation $\tilde{f}$ \citep[\eg][]{Bird.2010,Parno.2012,Regis.2014,Sun.2013}. Such approaches are also referred to as response surface method, surrogate model, or fitness function. However, in the aforementioned stream of research, the approximation is learned from the function calls to $f$ (\ie, surrogates are predicted based on past observations), while leaving the rest of the algorithm unchanged; the approximation is then merely used as a proxy for $f$, but any probabilistic interpretation does not further guide the search process. Hence, their only similarity with our approach is that these approaches also draw upon a function approximation, but for a completely different purpose.

Following the above rationale, surrogate-assisted PSO has been developed earlier \citep[\eg][]{sun2017surrogate, yu2018surrogate, wang2017committee}. Here algorithms directly model surrogates (\eg, sigmoid-like inertia weights in Modified PSO) or using function approximations as a proxy for $f$. In fact, (non-stochastic) function approximation has been shown to improve evolutionary algorithms, as the surrogate can be used to evaluate additional candidate solutions within a local neighborhood, while keeping the number of function calls to $f$ unchanged \citep{Regis.2014b}. Examples are surrogate-assisted variants of Gaussian PSO \citep[\eg][]{krohling2004gaussian, melo2016gaussian, varma2013gaussian, barman2016color, liu2013gaussian, gao2020novel}, Bayesian PSO \citep[\eg][]{zhang2015new, chen2017novel, kang2018optimal}, and modified PSO \citep[\eg][]{tian2018mpso, liu2015modified}. However, surrogate-assisted algorithms are primarily used to speed up runtime (due to fewer evaluations $f$) but with similar convergence characteristics. That is, the swarm movements are not directly adapted to the surrogate, whereas we use a surrogate to guide swarm search towards exploration-exploitation.  

\citet{Bird.2010} compute a local regression of the function $f$ around a certain point $x$ in order to yield an interpolation $\tilde{f}$ within a close neighborhood and then move the worst particle to the best location of the local regression. This approach resembles our algorithm~(B), but the specification in \citet{Bird.2010} is restricted to a local neighborhood and can thus take only a small region in consideration. Conversely, our Gaussian process regression computes the stochastic function approximation from all particles and thus yields a global stochastic function approximation. Thereby, it is able to relocate particles not locally but globally. In addition, \citet{Bird.2010} merely address local exploitation but ignore our idea of further exploration. 

Several attempt have been made to use surrogate models for directing a random search, yet outside of PSO. \citet{liu2013gaussian} propose a surrogate-assisted genetic algorithm based on Gaussian processes (GPEME for short). Here the surrogate is used for prescreening potential function evaluations, whereas our algorithm integrates the surrogate in way that it guides the search process (towards exploration-exploitation). There are further surrogate-assisted evolutionary algorithms for hierarchical swarm searches \citep{yu2018surrogate}, committee-based active learning approaches \citep{wang2017committee}, and  co-operative optimization \citep{sun2017surrogate}. However, different from the above, we develop a PSO variant on top of a Gaussian-process-based surrogate that leverages the uncertainty estimates in the GP, so that the swarm search is driven by both exploration and exploitation.

\section{PSO with Gaussian-process-based swarm search}
\label{sec:method}

\subsection{High-level description}
\label{sec:highlevel}

Our previous literature review has shown that particle swarm optimization suffers from returning local instead of global optima, as well as ignoring valuable information about the underlying functional landscape. Accordingly, our goal is to extend the strengths of this stochastic, population-based search method for black-box optimization problems with a more efficient search strategy. For this reason, we propose a combination of the PSO mechanism with a stochastic surrogate model of the objective function, so that the swarm search can be directed strategically. 

The stochastic surrogate model sheds light into the objective and we can leverage its landscape in the search process of PSO. Thereby, we aim at addressing two problems of traditional PSO algorithms, namely, the risk of finding local optima and redundant function evaluations, \ie, slow convergence in terms of evaluations. Our remedy to both is provided by learning a stochastic surrogate model: it can identify regions that have only barely been explored and where we thus face a high uncertainty regarding the landscape. On the basis of this, we can direct the swarm towards both exploration and exploitation. 

Key to the strategic search is an approximation of the underlying objective function. This is formulated as a learning task in which a stochastic model for the underlying function (surrogate model) is fitted. In this, we make use of ideas from the realm of Bayesian optimization to augment our particle swarm through a Gaussian-process-based surrogate model. The latter then guides the swarm towards regions that entail the optimum in our approximation of the objective, or where we have great uncertainty in our knowledge of the underlying function. The idea of a stochastic response surface is also the basis of Bayesian optimization methods. We give a review in \ref{appendix:bayesian_optimization}. We choose a Gaussian-process-based surrogate model for several reasons: (1)~Gaussian processes are common for modeling uncertainty in multivariate spaces \citep[cf.][]{ackermann2011nonlinear, Buche.2005}. The underlying Gaussian distribution makes few assumptions on the actual curvature of the function. (2)~Gaussian processes are fairly parsimonious (\ie, they have fewer model parameters as compared to many other surrogate models such as, \eg, artificial neural networks, support vector machines) and are thus beneficial when the swarm search has still little information. (3)~Gaussian distributions (that underlie GPs) are found to be highly effective in state-of-the-art Bayesian optimization \citep[cf.][]{Snoek.2012, swersky2013multi}.

%
%
In principle, the stochastic function model can facilitate swarm movements with regard to both exploration and exploitation, yet is is unclear whether it should influence purely directions, or even the velocity governing the step sizes. In the following, we thus suggest different approaches to how we integrate the Gaussian-process-based surrogate model into the search process and adapt the corresponding swarm intelligence:

\begin{enumerate}[label=(\Alph*)]
\item \textsc{PSO with heuristic direction}. Traditional implementations of particle swarm optimization move the walkers towards a mixture of personal and global best-known solutions. Conversely, we extend the velocity update and introduce a third direction that points towards the expected optimum from the approximation. This changes the velocity update fundamentally: while we previously merely utilized information from positions that we have already visited, the swarm could now even move towards directions that differ from the existing search trajectories. With each iteration, we augment our knowledge of the objective function and obtain an increasingly more accurate estimate of the landscape, including the approximate location of the corresponding optima. As a result, the swarm is additionally attracted by the optimum from the stochastic function model. The optimum of the stochastic function model is considered as an effective guess to the optimum of the function, and we use this information in combination with cognitive and social influence as a third \emph{heuristic influence} on the particles. 

\item \textsc{PSO with heuristic exploitation}. Often, an adjustment of the direction is not sufficient to accomplish a fast convergence behavior and we thus propose a variant where we directly relocate unsuccessful particles to the optimum in the approximation model. In other words, the particle is moved to the best average location given the stochastic function model. As a result, this accelerates convergence, as we explicitly query points which are likely to improve our swarm according to the stochastic function model, as opposed to relying on sheer luck. At the same time, the stochastic movement of the other particles give us continuous exploitation of the best we know. This corresponds to the combination of PSO with a response surface, albeit with a very general framework for response surfaces. 

\item \textsc{PSO with heuristic exploration}. The previous version ignores uncertainty in the stochastic function model for the most part, and thus does not have any incentive for exploration. For this reason, we utilize the stochastic nature of the GP model that allows us find uncertainty estimates for the stochastic function model. We further modify the exploration behavior of the swarm and, in each iteration, relocate one particle with the worst function value to a location with high uncertainty in the stochastic function model. As a result, this search strategy entails an explicit and formalized exploration process, which differs from the swarm movements in traditional variants of PSO. Mathematically, it also allows for a guaranteed convergence of the same complexity class as Bayesian optimization.

\end{enumerate}

In the following, we provide a definition of Gaussian processes as our function approximation and, based on this, we subsequently specify the actual optimization routines. 

\subsection{Gaussian processes}
\label{sec:gp}


A Gaussian process~(GP) over a domain $\domain$ is given by a mean function $m: \domain \to \R$ and a covariance function $K: \domain^2 \to \R$. The covariance function must define a valid covariance form, \ie it is a positive semi-definite kernel over the space $\domain^2$. Definiteness corresponds to non-degenerate GPs. Now let $\GP(m, K)$ denote the distribution given by the GP with mean $m$ and covariance $K$, which is given by the following definition. We refer to \citet{Rasmussen.2008} for definitions, posteriors and common choices of covariance functions. 

\begin{definition}[Gaussian Process]
\label{def:gp}
For any finite set of points $X = \{x_1, \dots x_n\}$, $f\sim \GP(m, K)$ implies
\begin{equation}
\label{eqn:gp}
f(X) \sim
\normal\left(
m(X),
K(X, X)
\right),
\end{equation}
where $f(X)$ is the vector with $f(X)_i=f(x_i)$, $m(X)$ the vector with $m(X)_i = m(x_i)$ and $K(X, X)$ is the matrix with entries $K(X, X)_{i, j} = K(x_i, x_j)$.
\end{definition} 

\begin{remark}[Gaussian Process Posterior]
	\label{rmk:gpprop}
Let $Y = \{y_1,\dots, y_m\}$ be another set of points, and let $K(X, Y)$ be the matrix with $K(X, Y)_{i, j} = K(x_i, y_j)$. The definitions of $K(Y, X)$ and $K(Y, Y)$ are analogous. Given an observation of $f(X)$, the posterior distribution over $f$ is then given by 
\begin{equation}
\label{eqn:gp_post}
\begin{split}
f(Y) \sim \normal \Bigl(&m(Y) + K(Y, X) K(X, X)^{-1}(f(X)-m(X))\\
&K(Y, Y) - K(Y, X) K(X, X)^{-1} K(X, Y) \Bigr)
\end{split}
\end{equation}
for any set $Y\subset\domain$. 
\end{remark}

As one can readily see, this means that the posterior distribution of a GP is again a GP with modified $m$ and $K$. Thus any posterior evaluation $f(Y)$ follows a multivariate normal distribution. We recall that the marginals of a multivariate normal distribution are also normal distributions. This allows us to efficiently calculate the distribution of function values of the GP model at arbitrary locations.

The choice we must make when specifying a GP as a stochastic surrogate model for our function are the covariance $K$, as well as the mean $m$. In practice, one sets $m=0$ and concentrates on adapting $K$ to the problem setting \citep[Chapter 4]{Rasmussen.2008}. This allows the user to encapsulate, a~priori, known properties of the function and we later give guidance concerning the choice. 

\subsection{Stochastic surrogate model through Gaussian processes}
\label{sec:stochastic_function_model}

We now define our stochastic surrogate model based on which we approximate the objective function. Let $\theta$ denote the additional parameters in a covariance $K_\theta$. Then, the interpolation requires one to identify suitable parameters $\theta$. Their values can be inferred from the observed data based on a maximum likelihood estimation
\begin{equation}
\label{eqn:theta}
\arg\max_{\theta} p(f(X) \mid \theta)
\end{equation}
using
\begin{equation}
p(f(X) \mid \theta) \propto \exp\left(-\frac{1}{2} f(X)\transpose K_\theta(X, X)^{-1} f(X) - \frac{1}{2}\log\det K_\theta(X, X)\right),
\label{eqn:kernfit}
\end{equation}
which gives the maximum a~posteriori estimates of $\theta$. The above optimization can be solved with a numerical optimization procedure of choice; \eg, quasi-Newton methods.\footnote{In our case, we use L-BFGS-B \citep{Byrd.1995} from the \Python\space package \texttt{scipy}. We used 10 restarts from random points on the parameter grid so as to ensure that we find a global optimum, or at least a sufficient local optimum.}

Even though this optimization problem is generally tractable, the matrix inversion in \Cref{eqn:kernfit} requires $\mathcal{O}(n^3)$ floating-point operations per evaluation. While this may not prove to be a bottleneck depending on the application, it does make a comparison over a wide variety of functions with multiple runs per function prohibitively expensive for our algorithms. This is why we use a greedy heuristic to only retain certain evaluations when fitting our stochastic surrogate model. That is, we use only a subset $\mathcal{M}\subset X$ of our observations, our \textit{memory}, and fit the model via  $\arg\max_{\theta} p(f(\mathcal{M}) \mid \theta)$. We describe the selection process in more detail in the following section.

\subsection{Efficient learning of stochastic surrogate model}

We only retain a subset of $\mathcal{M}\subset X$ of our observations. The corresponding selection is made by a function $\chi$ which extracts the point that are considered most informative. We define it in the following.

Recall $w_j^{(i)} := (x_j^{(i)}, v_j^{(i)}, p_j^{(i)})$ is the particle $j$ at step $i$ (\cf \Cref{sec:PSO}). We write $\particles{i}$ for the set of particles at step $i$, $\memory{i}=\{y_1,\dots, y_k\}$ for the set of observations we keep in memory, and use the notation $f(\particles{i}) = (f(\position{1}{i}),\dots, \position{{\npart}}{i})$ as for $f(X)$ in \Cref{sec:gp}. Further, we write $f(\memory{i}\cup\particles{i}) = (f(\position{1}{i}), \dots, f(\position{{\npart}}{i}), f(y_1), \dots, f(y_k))$. 

As outlined in the previous section, we use a greedy heuristic to only retain a subset of the locations and evaluations at each iteration in $\memory{i}$.
At each time step, we use the current stochastic surrogate model $\GP(m, K_\theta)$ and set 
\begin{equation}
\begin{split}
\chi(m, K_\theta, \particles{i}) = \Bigl\{\position{j}{i} \;\big|\; f(\position{j}{i}) \notin \Bigl[\;&m(\position{j}{i}) - \rho \, K_\theta\left(\position{j}{i},\position{j}{i}\right)\\&m(\position{j}{i}) + \rho \, K_\theta\left(\position{j}{i},\position{j}{i}\right)\;\Bigr]\Bigr\}
\end{split}
\label{eqn:heuristicChoice}
\end{equation}
for a fixed value of $\rho$. We set $\rho$ consistently to correspond to the $75\,\%$ confidence interval, \ie $\rho\approx1.15$. This heuristic corresponds to only keeping informative observations, that is, observations which change our assumptions about the function. All observations which appear likely under the current stochastic surrogate model are forgotten. We additionally keep all most recent observations in memory, to ensure that our stochastic function model is precise near our swarm.

\subsection{Optimization methods}
\label{sec:methods}

We now formalize the different optimization methods combining both PSO and the Gaussian-process-based surrogate model. To this end, we first specify the general form that is shared across all three algorithms (see \Cref{alg:base}) and later discuss the individual modifications. The different algorithms only differ in the update rule that is given \Cref{ln:update} of \Cref{alg:base}, where each variant inserts its own rule. The general layout of \Cref{alg:base} first consists of initializations, where the initial particle positions are determined and the initial set of points for the stochastic function approximation are set. A subsequent loop moves the swarm around the function space. Within this loop, we fit the stochastic function model in \Cref{ln:paramfit,ln:gpfit}. Then, \Cref{ln:update} moves the particle around, while \Cref{ln:memory} updates the memory that underlies our heuristic for accelerating the learning of the GP-based surrogate model. 

\begin{algorithm}
	\caption{General form of PSO with a Gaussian-process-based surrogate model.}
\footnotesize
\begin{algorithmic}[1]
\footnotesize
	\REQUIRE Domain $\domain$, parameterized family of kernels $K_\theta$, function $f$
	\ENSURE Approximate global minimum $x^\ast$ of $f$
	\STATE Initialize set of $\npart$ particles $\particles{0}$ uniformly at random over $\mathcal{D}$ with multivariate normal velocity
	\STATE Set $\memory{0} \leftarrow \particles{0}$
	\FOR{$i = 1$ to convergence}
		\STATE Fit $\hat{\theta}$ via maximum likelihood of $f(\memory{i-1}\cup \particles{i-1}) \;\Big|\; \GP(m, K_\theta)$ \label{ln:paramfit}
		\STATE Compute stochastic function approximation $\GP(\tilde{m}, \tilde{K}_{\hat{\theta}})$ as the posterior of $\GP(m, K_{\hat{\theta}})$ for the observations $f(\memory{i-1}\cup\particles{i-1})$ \label{ln:gpfit}
		\STATE Move particles via $\particles{i}\leftarrow$ update$\left(\tilde{m}, \tilde{K}_{\hat{\theta}}, \mathcal{W}^{(i)}\right) $, where the update rule depends on the algorithm variant \label{ln:update}
		\STATE Updated memory via $\memory{i}\leftarrow \memory{i-1}\cup \chi(\tilde{m}, \tilde{K}_\theta, \particles{i-1})$ \label{ln:memory}
	\ENDFOR
	\RETURN Global best solution from $\particles{\text{end}}$
		\end{algorithmic}
	\label{alg:base}
\end{algorithm}

We now give the update rules corresponding to the individual algorithm variants:
\begin{enumerate}[label=(\Alph*)]
	\item \textsc{PSO with heuristic direction}. This algorithm extends the traditional velocity update by an additional direction towards the optimum from the stochastic surrogate model. More precisely, we take the mean of the Gaussian process as a response surface, and use the minimum of this response surface as a heuristic direction, \ie $h^{(i)} = \argmin_{x \in \domain} m(x)$. We then set the heuristic influence (HI) of particle $j$ to $h^{(i)} - \position{j}{i}$. We then use the update rule 
	\begin{equation}
	\velocity{j}{i+1} = \omega \velocity{j}{i} + \phi_p R_p \odot \text{CI} + \phi_g R_g \odot \text{SI} + \phi_h  R_h \odot \text{HI}, 
	\end{equation}
	in the SPSO algorithm. As a result, the convergence analysis of SPSO found in \citet{Jiang.2007} readily extends to this form, assuming convergence of the heuristic minimum $h^{(i)}$. \\
	 The question remains as to how we wish to weight the factors $\phi_p$, $\phi_g$ and $\phi_h$ in relation to each other. We evaluate three settings (A1)--(A3) with different parameters. For all other parameter, we use the intuition from SPSO to set the parameters of the PSO to sensible values.\footnote{In practice, this means that we retain $\omega$ as in SPSO, and let $\phi_p + \phi_g + \phi_h \approx \psi_p + \psi_g$, where $\psi_p$ and $\psi_g$ are the corresponding cognitive and social weights from SPSO.} 
	
	\item \textsc{PSO with heuristic exploitation}. This algorithm accelerates exploitation, as the \textquote{worst} particle in the swarm moves to the optimum of the stochastic function approximation. All other particles move according to the previous update rules. At each step, we choose the worst particle ${j^\ast} \in \argmin_{j=1, \dots, \npart} f(\position{j}{i})$ and determine its new location from the GP mean, \ie $x^\ast \argmin_{x \in \domain} m(x)$. Its new location and velocity is initialized via
	\begin{align}
	\position{{j^\ast}}{i+1} &= x^\ast,\\
	\velocity{{j^\ast}}{i+1} &\sim \normal(0, 1) \; \text{componentwise}, \\
	\personal{{j^\ast}}{i+1} &=\begin{cases}
	\personal{{j^\ast}}{i}, \quad &\text{if} \; f(x^\ast) > f(\personal{{j^\ast}}{i}),\\
	x^\ast, \quad & \text{otherwise}.
	\end{cases}
	\end{align}
	
	\item \textsc{PSO with heuristic exploration}. Here we use a procedure similar to version~(B), but now choose the new location according to a criterion encapsulating the uncertainty of the function value at this point. The only difference appears in our choice of the location $x^\ast$. We introduce an acquisition function $R$ which incorporates the posterior variance $\sigma(x^\ast):=K(x^\ast, x^\ast)$. The new location is then chosen by $x^\ast\in\argmax_{x\in\domain} R(K, m, x)$. Here we experiment with two choices of $R$. In version~(C1), we let the worst particle move to the optimum of the lower bound of the $90\,\%$ confidence interval, that is $R(K, m, x) = -(m(x) - \rho \sigma(x))$ for $\rho\approx1.6$. In the other version~(C2), we move it to the position of maximal uncertainty, that is use $R(K, m, x) = \sigma(x)$. Both versions perform added exploration, either by explicitly querying high-uncertainty points, or by being very optimistic as to the location value. \\
	The greedy heuristic in the stochastic function approximation makes a direct transfer of convergence guarantees difficult. However, if we include all past values in our Gaussian process, we can guarantee the same convergence behavior as the associated Bayesian optimization scheme, albeit with a higher number of concurrent function evaluations.
	
	\end{enumerate}

Each of the algorithms (A)--(C) relies internally on finding a new location from the stochastic function approximation. We note here that the identification is computationally efficient, as the location is either analytically known or one can use numerical optimization (\eg quasi-Newton methods). Multiple restarts are replaced by initializing the search with the location of the current optimum. 

\subsection{Choice of covariance function}

We consistently chose the covariance matrix
\begin{equation}
K(x, y) = \alpha_1^2 \exp \left(- \frac{\norm{x-y}^2}{\rho^2} \right) + \alpha_2^2 + \alpha_3^2 \,\indicator\{x=y\},
\label{eq:kernel}
\end{equation}
where the parameters $\alpha_1$, $\alpha_2$, $\alpha_3$ and $\rho$ are adapted by the estimation process. This choice of covariance corresponds to the assumption that our function has the form $f(x) + m + \varepsilon$, where $m$ is a global mean, $f$ is a smooth function with mean $0$, and $\varepsilon$ is white noise \iid at every point. This residual white noise is introduced to account for non-smoothness in the underlying function.

\subsection{Illustrative example}
\label{sec:example}


We give a short example of the proposed algorithm. For this, we look at several snapshots of the algorithm running on the two-dimensional Ackley function given by
\begin{equation}
\begin{split}
f(x, y) = &-20 \exp\left( -0.2 \sqrt{0.5 (x^2 + y^2)}\right)\\ & - \exp\left(0.5(\cos 2 \pi x + \cos 2 \pi y)\right) + e + 20,
	\end{split}
\label{eqn:ackley}
\end{equation}
with $e$ as Euler's number. The domain is set to $\intervalcc{-5}{5}^2$, for which the function has a global minimum of $0$ at $x=y=0$.

\begin{figure}[h]
\centering
\tiny
\makebox[\textwidth]{%
\begin{tabular}{p{4cm}p{4cm}p{4cm}p{4cm}}
\hspace{2.3cm}\mcellt{Objective\\function} & \hspace{1.5cm}\mcellt{Approximation\\(mean)} & \hspace{1.cm}\mcellt{Approximation\\(variance)} & \hspace{.6cm}\mcellt{Acquisition\\function} \\
 \multicolumn{4}{c}{\includegraphics[width=\textwidth]{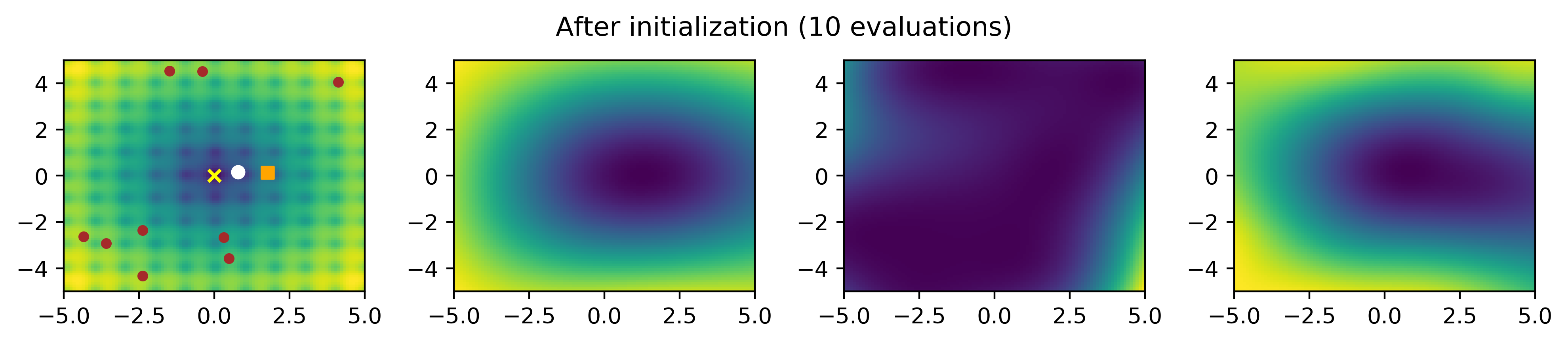}} \\
\multicolumn{4}{c}{\includegraphics[width=\textwidth]{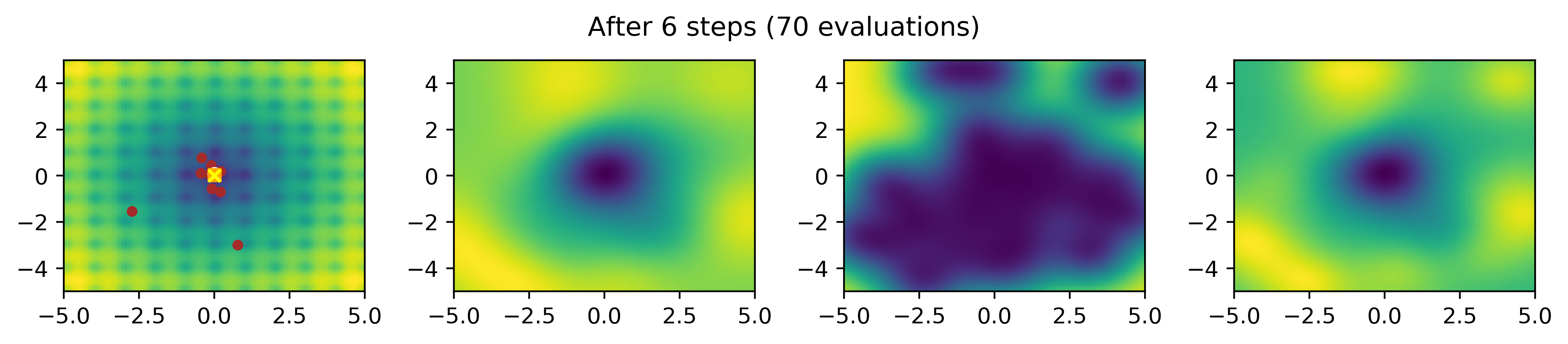}} \\
\multicolumn{4}{c}{\includegraphics[width=\textwidth]{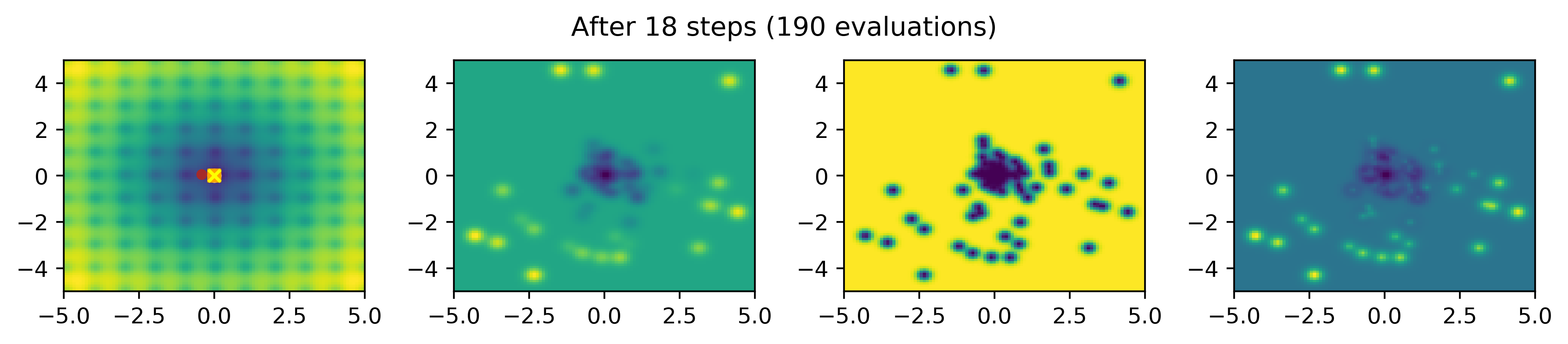}}
\end{tabular}
}
	\caption{Illustrative execution of the PSO with heuristic exploration, \ie variant~(C1). Shown is the state of the algorithm after initialization, 6 steps and 18 steps. Highlighted are the global optimum (yellow cross), the current heuristic exploration (white circle), the current best solution (orange square) and the particles (brown dots). The algorithm quickly learns a crisp approximation of the true objective and, based on it, guides the search direction of the particle swarm. Darker colors correspond to lower values. After 18 steps, almost all particles have converged to the true optimum and are so closely scattered around $x = y = 0$ that one can barely recognize them any longer.}
  \label{fig:examplerun}
\end{figure}

As an illustrative example, we now execute PSO with heuristic exploration, more precisely, variant~(C1) of our algorithm. We use 10 walkers and otherwise the same parameters as in \Cref{sec:results}. \Cref{fig:examplerun} depicts different iterations of the algorithm, namely, after initialization, after 6 steps (\ie 70 function evaluations) and after 18 steps (\ie 190 function evaluations). After initialization, the stochastic surrogate model has learned little about the true nature of the wavy objective function and, hence, its mean suggests a rather smooth curve with a high variance associated to it. The acquisition function shows the lower bound of the $90\,\%$ confidence interval. Accordingly, the heuristic exploration guides the algorithm to further explore areas that promise both high chances of improvements, while taking the point-specific uncertainty into consideration.  

As the algorithm progresses, the particles quickly converge towards the optimum in the center of plot, while the mean surface yields a better resolution and serves as an increasingly accurate function approximation. Hence, the proposed location from the heuristic exploration quickly coincides with the true optimum. We can also see the convergence of the particles towards the global optimum of the Ackley function at $x = y = 0$. After 18 steps, we can barely recognize the position of individual particles as these are so closely scattered around the optimum. Here we also note the low variance of the approximation around the optimum. Similarly, the acquisition function also attains its minimum around $x = y = 0$ and thus guides the swarm to explore this area.

\FloatBarrier
\section{Computational experiments}
\label{sec:results}

\subsection{Setup}
\label{sec:setup}

We performed an extensive series of computational experiments. For reasons of comparability, we adhere to the CEC2013 benchmark functions and adopt the same experimental setup as in \citet{liang2013problem}. The benchmark functions are reviewed in \ref{appendix:cec2013}. These are challenging in different ways (\eg, their landscape suffers from multi-modality and poor scaling). 


In accordance with literature \citep{Bonyadi.2017}, we experiment with SPSO2011 as the prime benchmark algorithm. In our experiments, we have set the parameters of SPSO2011 according to the suggestions in \citet{ZambranoBigiarini.2013} for reasons of comparability (details in \Cref{tab:algos}). All PSO variants are compared with 50 particles \citep{yu2018surrogate}. In practice, PSO is often used in settings where function evaluations are computationally or economically costly \citep{Rios.2013}. This is because, for instance, complex industrial simulations are involved \citep[\eg][]{hong2018optimal}. This is the setting to which our algorithms are tailored, and, hence, our evaluations compare the efficiency of the algorithms after $100 \, D$ function evaluations following \citet{liu2013gaussian, varma2013gaussian}. For reasons of space, we report the results for ten dimensional ($D=10$) objective functions. In addition, we utilize Bayesian optimization as another comparison, as our approach borrows concepts from the underlying GP-based approximation of functions, though the stochastic process in PSO yields a completely different search strategy. We evaluated Bayesian optimization in the standard parallelized implementation of the \Python\space package \texttt{GPyOpt}, which uses local penalization \citep{Gonzalez.2015}. For each experiment, we measure the performance regarding the mean, minimum, maximum, median, and the standard deviation across 51 independent runs as in \citep{liang2013problem, ZambranoBigiarini.2013}. 

\begin{table}
	\centering
	\footnotesize
	\renewcommand{\arraystretch}{0.618}
	\makebox[\textwidth]{
	\begin{tabular}{ll}
	  \toprule
		\textbf{Algorithm} & \textbf{Parameters} \\
		\midrule
		Version~(A1) & $\omega = 0.42, \phi_p=1.2, \phi_g=1.2, \phi_h=0.75$\\[2pt]
		Version~(A2) & $\omega = 0.42, \phi_p=1.55, \phi_g=0.75, \phi_h=0.75$\\[2pt]
		Version~(A3) & $\omega = 0.42, \phi_p=0.75, \phi_g=1.55, \phi_h=0.75$\\[2pt]
		Version~(B) & $\omega = 0.42, \phi_p=1.55, \phi_g=1.55$\\[2pt]
		Version~(C1) & $\omega = 0.42, \phi_p=1.55, \phi_g=1.55, R(m, K, x)=m(x)-1.6\sigma(x)$\\[2pt]
		Version~(C2) & $\omega=0.42, \phi_1=1.55, \phi_g=1.55, R(m, K, x)=-\sigma(x)$ \\[2pt]
		SPSO2011 & $\omega = \frac{1}{2  \ln{2}}$, $\phi_p = 0.5 + \ln(2), \phi_g=0.5 + \ln(2)$\\[2pt]
		Bayesian optimization & Default parameters as in \texttt{GPyOpt} library\\
		\bottomrule
	\end{tabular}
  }
	\caption{Overview of the evaluated algorithms where (A)~refers to our PSO variant with a GP-based heuristic; (B)~is the exploitation variant; and (C)~is the exploration variant.}
	\label{tab:algos}
\end{table}

\FloatBarrier
\subsection{Numerical performance}

The results of the computational experiments are provided in \Cref{tab:computationalresultsA1A2A3B,tab:computationalresultsC1C2SPSO2011BO}. Overall, the variants of our proposed algorithm outperforms SPSO2011 and Bayesian optimization (BO) in all 28 experiments. The improvements are as large as 64\,\%. In sum, he performance of the proposed algorithm strongly exceeds the performance of previous implementations of PSO.


We now discuss the performance of different variants of our proposed search strategy. Overall, our evaluation favors variant~(A3), which regularly outperforms the other variants (\ie, in 13 out of 28 experiments), followed by variant~(B) (7 out of 28 experiments). The favored variant~(A3) improves the performance of SPSO2011 for multimodal experiments (\eg, f15) by up to 64\,\% and for composition functions (\eg, f23) by up to 36\,\%. Furthermore, variant~(A3) outperforms both the other variants of our algorithms and SPSO2011 for multimodal functions in 9 of 15 experiments. For composition functions, we observe a similar result. Here variant~(A3) achieves the best overall performance in 4 out of 8 experiments. Therefore, the explorative implementation of PSO with heuristic direction (A3) drives the search more effectively to the minimum of complex functions. 

In comparison, variants~(A1) and (A2) are each superior in only a single experiment. Both variants~(C1) and (C2) outperform the remaining alternatives of our algorithms and benchmarks in 3 experiments. Recall that variant~(A) of our PSO introduces a heuristic direction as a third proponent for the swarm. Here we vary the weights of the cognitive, social, and heuristic influence. Variant~(B) comprises a heuristic exploitation inside the PSO algorithm, while variant~(C) accentuates exploration, either with regard to a lower confidence bound~(C1) or the highest uncertainty~(C2). In sum, heuristic exploitation in the PSO algorithm facilitates the optimization of complex functions as compared to the heuristic exploration. We obtain the overall best results when drawing upon a heuristic direction in the PSO algorithm. 

We further compare the performance through statistical tests (see \ref{appendix:stattests}). That is, we compare the performance of the favored variant~(A3) against the performance of benchmark algorithms to see whether the improvement is statistically significant. For a significance threshold of $\alpha=5\,\%$, variant~(A3) is superior over SPSO2011 for 27 of the 28 objective functions at a statistically significant level. This confirms the effectiveness of the proposed algorithms.

\begin{table}[H]
    \tiny
    \thisfloatpagestyle{empty}
    \centering
    \setlength{\tabcolsep}{3pt} 
    \renewcommand{\arraystretch}{0.85} 
    \centerline{
    \begin{tabular}{>{}l>{}r|>{}r>{}r>{}r>{}r>{}r|>{}r>{}r>{}r>{}r>{}r}
            \toprule
            \textbf{Function} & $\bm{f(x^*)}$ & \textbf{Min} & \textbf{Median} & \textbf{Mean} & \textbf{Max} & \textbf{SD} & \textbf{Min} & \textbf{Median} & \textbf{Mean} & \textbf{Max} & \textbf{SD} \\
            \midrule
            \multicolumn{7}{c|}{\textbf{Variant (A1)}}   &  \multicolumn{5}{c}{\textbf{Variant (A2)}} \\\midrule
            \multicolumn{3}{l}{\textsc{Unimodal functions}}  \\[2pt]
            f1 & -1400 & {-1397.5872} & {-1383.4696} & {-1361.6633} & {-1043.4621}{}  & {63.3769} 
            &  -1396.0733 &  -1341.5089 & -611.4317  &  1.2080e+05  &  2377.0420 \\
            f2 & -1300 & 2.6277e+06  & 2.5977e+07  & 3.9917e+07  &  2.2339e+08  & 4.7645e+07  
               & 7.4935e+06  &  5.2075e+07 &  1.0334e+08 &  4.1734e+08  &  1.0360e+08 \\
            f3 & -1200 & \textbf{4.1720e+06}  & \textbf{1.3255e+10}  & \textbf{5.2153e+10}  &  \textbf{1.2927e+12}  & \textbf{1.8859e+11}
               &  2.1073e+09 &  2.8896e+10 & 8.3549e+11  &   1.3209e+13  & 2.7495e+12  \\
            f4 & -1100  & 1.4708e+04 & 5.3500e+04 &  8.7354e+04 &  6.2230e+05 & 1.2092e+05 
               & \textbf{4766.9000}  & \textbf{5.4850e+04}  & \textbf{2.3056e+05}  &  \textbf{3.7958e+06}  & \textbf{5.9561e+05}\\
            f5 & -1000  & -951.5490 & -70.1066  & 662.2853  &   1.2127e+05 &   2238.8980 
               &  -619.4898 & 1419.3945  &  3961.1237 &   3.178e+05 &  6624.0905 \\[5pt]
           \multicolumn{3}{l}{\textsc{Multimodal functions}} \\[2pt]
            f6 & -900   & {-893.0991}  & {-802.8364}  & {-782.0918}  &  {-133.5772}  &  {112.9641}
               & -866.3261  & -697.4170  &   -603.4517 &  451.9740  & 268.6724  \\
            f7 & -800   & -768.6168 & -625.5504  & -494.6803  &  2049.3854  &  475.6174
               & -722.3400  & -541.1433  &  -117.2380 &  3836.2694  &   993.0082 \\
            f8 & -700   & -679.3985  &  -679.0972 & -679.0835  &  -678.7793  &   0.1434
               & -679.3306  & -679.0104  &  -679.0234 &  -678.8371  &  0.1278 \\
            f9 & -600   &  {-597.7262} &  {-591.5365}  & {-591.9679}  & {-585.6585}   & {3.1091}
               & -591.7658  & -590.7448  & -590.7448  & -589.7239   &  1.4439 \\
            f10 & -500   & -487.8997  &  -412.2499 & -292.6228  & 863.0087   & 277.7393  & -480.5477  &  -303.1307 &  -81.0474 &   1630.5215 &  489.4571 \\
            f11 & -400   & -361.5090  &  -336.4743 & -334.7380  &  -289.1876  &  14.7983 & -365.4087  & -320.4254  &   -311.2982 &  -139.7896  &  37.1877 \\
            f12 & -300   &  -265.5353  & -227.4410  & -226.6150  &  -157.8276  &  20.0157 & -261.8385  & -204.3804  & -196.7491  &   -76.1679  & 37.4749  \\
            f13 & -200   &  -164.2729  &  -129.9400 & -127.4905  &  -76.9984  &  18.3428 &  -148.0190 &  -110.7818 & -104.1624  &   21.3197  &  32.7836 \\
            f14 & -100   &  525.3739  &  1478.9113 & 1474.5626  &    2548.0959  &  457.6333  
                & 503.9011  &  1767.5706 & 1837.6050  &  3006.6952  &  561.8864 \\
            f15 & 100   & {945.4400}  & {1979.8693}  &  {1925.1920} & {3042.6037}   &  {440.3659}  
                & 1177.2460  & 2150.8842  &  2111.5859 &  3060.6540  &  487.3763 \\
            f16 & 200   &  200.9729 & 203.0626  &   203.0612 &   204.8510  &   0.8668  
                & 201.3839  & 203.6982  & 203.5232  &  206.5214  &  1.1423  \\
            f17 & 300   &  {300.0980}  & {302.1162}  & {307.6036} &  {408.6987}  &  {17.6225}  
                & 300.2047  & 304.7179  &  317.5512 &  543.9282  &  38.8794 \\
            f18 & 400   & 400.1664  &  402.5121 &  405.8163  &  443.5966 &  7.9026  
                & 400.2131  & 408.2567  &  421.2932  &  606.5287  &  39.6921  \\
            f19 & 500   & {503.0887}  &  {515.4676} &  {1339.9376} &  2.2403e+05 & {3400.5543} 
                & 505.4403  &  2596.9577 & 4.2600e+05 &   5.5700e+06 &  9.0404e+05 \\
            f20 & 600   & {603.6799}  & {604.6531}  &   {604.6227} & {605.0000}   &   {0.3463} 
                & 604.1277  &  604.8752 & 604.7817 &  605.0000  &  0.2501 \\[5pt]
            \multicolumn{3}{l}{\textsc{Composition functions}} \\[2pt]
            f21 & 700   & 1196.3619  & 1527.3981  & 1543.4502  &  2760.8438   &  200.5887  
                & 1244.2648  & 1614.4777  & 1756.0408 &   2656.2074  & 329.0863  \\
            f22 & 800   &  1602.6127  &  2694.7480  &  2676.1522 &  3733.3636  & 477.6080   
                & 1838.6323  &  2986.3565 &  2953.3486 & 3908.0153    &  502.4572 \\
            f23 & 900   & {1834.8290}  & {3024.2683}  &  {2936.5531}  & {4121.2138}   & {510.9749}  
                &  1952.3371  & 3108.7894  & 3131.3746  &  3953.9239  &  460.8704 \\
            f24 & 1000   &  1152.8172 & 1228.8007  &  1224.1155  &  1280.8363  &   24.8739 
                &  1179.1119  & 1239.1245  & 1238.7175  &  1331.0587  &  24.9041 \\
            f25 & 1100   & 1256.1385  & 1350.9986  &   1344.9651 & 1376.9652   &   28.1815 
                & 1262.6632  & 1359.1513  &  1358.0976 & 1382.6490   &  20.4727  \\
            f26 & 1200   & 1348.2933  &  1403.5477  &  1424.4960  &  1548.7305  &  57.6909
                &  1363.1698  & 1418.9811  & 1437.5508  &  1547.1016 &  50.8241 \\
            f27 & 1300   & 1636.9476   & 1654.0901  &  1655.2239 &  1705.1864  &  11.3101  
                & 1642.7216  & 1660.8510  & 1668.7147  &   1817.8577 & 26.4306  \\
            f28 & 1400   & {1543.1449} & {2171.8226}  & {2135.7050}  &  {2811.7222}  & {233.6992}
                & 1625.6656  & 2409.6956  & 2526.8903  &  4038.9449  &  452.7039 \\\midrule
            \multicolumn{7}{c|}{\textbf{Variant (A3)}}   &  \multicolumn{5}{c}{\textbf{Variant (B)}}\\\midrule
            \multicolumn{3}{l}{\textsc{Unimodal functions}} \\[2pt]
            f1 & -1400 & \textbf{-1399.8136} & \textbf{-1396.7569}  & \textbf{-1395.0071}  &  \textbf{-1372.3346}  & \textbf{5.6353}  
                       & {-1397.5894}  & {-1376.2295}  & {-1338.3509}  &  {-677.2732} & {109.2581} \\
            f2 & -1300 &  1.0587e+06  &  7.9735e+06 & 1.2012e+07  &  7.7522e+07  &  1.2741e+07 
               & {9.7927e+05}  &  {1.0826e+07} &  {1.5301e+07} &   {7.3415e+07}  &  {1.5404e+07} \\
            f3 & -1200 &  1.5196e+06 & 7.1068e+08  &  2.3204e+09 &  1.7155e+10  & 4.1159e+09
               & 1.7493e+07  &  {9.7294e+08} & 2.8782e+09  &   2.0762e+10  &  4.4020e+09 \\
            f4 & -1100  & 7876.3791  &  4.0557e+05  & 4.0105e+05 &  7.8208e+05 &   1.5994e+05
               &  8604.8093  &  4.5064e+05  &  5.3138e+05 &  3.7933e+06 &  5.0677e+05 \\
            f5 & -1000  & -930.3052  & -549.1494  & -381.1171  & 1582.2375   & 556.9127
               & -994.8325  & {-884.9113}  &  -766.3978 & 822.9534   &  314.2315 \\[5pt]
            \multicolumn{3}{l}{\textsc{Multimodal functions}} \\[2pt]
            f6 & -900   &  \textbf{-897.4271}  &  \textbf{-860.0267} &  \textbf{-848.4268} &  \textbf{-767.2946}   &  \textbf{37.2299}
               &  -888.4831  & -820.1558  & -823.0307  &  -682.8588  &  49.9664 \\
            f7 & -800   & -772.9237 & -691.0189  &  -684.6283  &  -574.2905  &   45.7213
               & -782.9525  & -683.7158  & -682.9493  &   -467.0235  & 56.9393  \\
            f8 & -700   &  -679.4831  & -679.1640  & -679.1858  &  -678.9770  &  0.1205
               & -679.5066  & -679.1245  & -679.1450  &  -678.9398  &   0.1396 \\
            f9 & -600    &  \textbf{-598.4306} & \textbf{-594.8041}  &  \textbf{-594.4478} &   \textbf{-589.2729}  &  \textbf{2.2898}
               & -591.7658  &  -590.7448 &  -590.7448 &   -589.7239  &  1.4439  \\
            f10 & -500   &  \textbf{-498.2982} &  \textbf{-474.0979} & \textbf{-451.9723}  &  \textbf{-262.2669}  & \textbf{54.8691}
                & -494.7935  &  -448.2590 & -419.9692  &  136.3982  &  112.7522 \\
            f11 & -400   & -378.8572  &  -356.2611 &  -355.2700 &  -330.1124  &   10.8450
                & \textbf{-384.9150}  & \textbf{-348.6396}  & \textbf{-347.7882}  &  \textbf{-310.3917}  &  \textbf{18.1133} \\
            f12 & -300   &  -265.9925 & -248.3478  & -247.0188  &   -226.2649 &   10.2883 
                & \textbf{-272.1120}  &  \textbf{-241.0569} &  \textbf{-235.7232} &  \textbf{-129.0170}  &  \textbf{25.8599} \\
            f13 & -200   &  -178.7481 &  -148.8482 & -146.9601  &  -121.7209  &  12.9224
                & \textbf{-179.8161}  &  \textbf{-135.8240} & \textbf{-133.9180}  &  \textbf{-85.5803}  &  \textbf{18.6451}  \\
            f14 & -100   & \textbf{231.1933}  & \textbf{1032.8714}  & \textbf{1086.9536}  &  \textbf{1883.7150}  & \textbf{489.8845}
                &  552.2438 &  1436.0180 &  1420.6820 &   2274.5364  &  442.6276 \\
            f15 & 100   &  \textbf{699.5687} & \textbf{1612.5707}  & \textbf{1615.8668}  &  \textbf{2413.8637}  & \textbf{444.6132} 
                & 1215.4654  & 1955.1194  & 1939.6750  &  2652.7653  &  369.7568 \\
            f16 & 200   &  200.8584 & 202.5599  &  202.4624 &  204.3681  &  0.7714
                &  \textbf{200.8200} & \textbf{202.8882}  & \textbf{202.7334}  &  \textbf{205.1360}  &  \textbf{0.8536}  \\
            f17 & 300   & \textbf{300.0100}  & \textbf{300.2201}  & \textbf{301.8915}  &  \textbf{326.4162}  & \textbf{ 5.0410}
                & 300.1950  &  301.0488 &  304.4055 &  344.4286  &  9.3396 \\
            f18 & 400   &  \textbf{400.0146} & \textbf{400.2460}  &  \textbf{401.501} &  \textbf{420.4141}  &  \textbf{3.5715}
                & 400.1594  & 401.0141  &  402.3797  & 426.3327   & 4.1385  \\
            f19 & 500   &  \textbf{501.6649}  & \textbf{504.9664}  & \textbf{507.3177}  &  \textbf{561.5510}  &  \textbf{8.8034}
                & 503.2288  & 506.0481  & 506.3795  &   513.1465  &  1.9141  \\
            f20 & 600   &  \textbf{603.1930} &  \textbf{604.3008}  & \textbf{604.2748}  &  \textbf{605.0000}  &  \textbf{0.4583}
                &  603.7083 & 604.3799  &  604.3923 &  605.0000  &  0.3440  \\[5pt]
            \multicolumn{3}{l}{\textsc{Composition functions}} \\[2pt]
            f21 & 700   & \textbf{1033.4041}  & \textbf{1512.7487}  & \textbf{1473.8847}  &  \textbf{1552.7165}  & \textbf{120.6318}
                & {1075.0764}  & {1512.8754}  & {1399.4236}  &  {1521.6803}  &  {172.1156}  \\
            f22 & 800   &  1456.5905 &  2257.5447 &   2247.2375 &  3120.9868  &  368.3120
                &  {1499.1383} & {2590.8005}  & {2468.9952}  &  {3454.4903}  & {548.5719}  \\
            f23 & 900   &  \textbf{1825.3987} & \textbf{2600.1388}  & \textbf{2646.5041}  &  \textbf{3690.7404}  &  \textbf{472.2147}
                & 1969.5557  &  2774.3180  &  2849.9366 &  3892.0847  & 395.0898  \\
            f24 & 1000   &    1133.6772 &  1219.6431 & 1210.1116  &  1242.1983  &  28.5619
                & \textbf{1121.1464}  & \textbf{1225.0830}  & \textbf{1218.1478}  &  \textbf{1243.7913}  &  \textbf{26.2537} \\
            f25 & 1100   &  \textbf{1232.3341} & \textbf{1342.3835}  &   \textbf{1334.7156} &  \textbf{1365.1327}  &  \textbf{29.6783}
                & {1241.8700}  & {1345.7829}  &  {1340.0374} &  {1366.6899}  & {24.7427} \\
            f26 & 1200   & 1331.7175  &  1400.6005 & 1409.8450  &  1529.4555  & 53.9006
                &  \textbf{1326.0919}  & \textbf{1400.6331} & \textbf{1409.5518}  &  \textbf{1533.1112}  &  \textbf{51.6916}  \\
            f27 & 1300   &  1626.7491 & 1640.0832  &  1641.3110 &  1666.9729  &   8.1147
                & \textbf{1623.8144}  &  \textbf{1643.4403} & \textbf{1643.2523}  &  \textbf{1665.1437}  & \textbf{7.3174}  \\
            f28 & 1400   & 1515.9583  &  2141.2106  &  2043.2591  &  2265.9658  &  225.6466
                & 1549.1988  & 2141.2374  &  1947.3751 &  2151.6891  & 344.8704  \\
            \bottomrule
            \end{tabular}}
		    \caption{Performance of PSO with a GP-based heuristic on CEC2013 benchmark experiments. Bold: Variant with lowest min value.}
            \label{tab:computationalresultsA1A2A3B}
\end{table}

\begin{table}[H]
    \tiny
    \centering
    \thisfloatpagestyle{empty}
    \setlength{\tabcolsep}{3pt} 
    \renewcommand{\arraystretch}{0.85} 
    \centerline{
    \begin{tabular}{>{}l>{}r|>{}r>{}r>{}r>{}r>{}r|>{}r>{}r>{}r>{}r>{}r}
            \toprule
            \textbf{Function} & $\bm{f(x^*)}$ & \textbf{Min} & \textbf{Median} & \textbf{Mean} & \textbf{Max} & \textbf{SD} & \textbf{Min} & \textbf{Median} & \textbf{Mean} & \textbf{Max} & \textbf{SD} \\
            \midrule
            \multicolumn{7}{c|}{\textbf{Variant (C1)}}   &  \multicolumn{5}{c}{\textbf{Variant (C2)}}\\\midrule
            \multicolumn{3}{l}{\textsc{Unimodal functions}} \\[2pt]
            f1 & -1400 & -1395.6798  & -1371.9866  & -1338.8192  &   -665.6236  &  114.7387 
               & -1396.1872 &  -1371.9617 & -1355.3548  &  -1229.9828  & 42.2859  \\
            f2 & -1300 &  \textbf{9.6679e+05} & \textbf{1.0576e+07}  &  \textbf{1.5863e+07}  &  \textbf{8.7980e+07}  & \textbf{1.7197e+07 }
               & 9.6679e+05  &  1.0774e+07  & 1.5359e+07  &  7.3415e+07  &  1.5541e+07 \\
            f3 & -1200 & 1.7493e+07  &  9.7295e+08  &   2.8782e+09 &   2.0762e+10  &   4.4020e+09
               & 1.7493e+07  & 9.9883e+08  & 2.9884e+09  & 2.0762e+10   &   4.6035e+09 \\
            f4 & -1100  & 8604.8093  & 4.5064e+05 & 5.3009e+05 &  3.7933e+06 & 5.0720e+05
               & 8604.8093  &  4.2507e+05 & 5.2690e+05 &  3.7933e+06 & 6.3108e+05 \\
            f5 & -1000  & {-994.9976} & {-865.5925}  &  {-740.5261} & {1458.8710} &  {396.5875}
               & \textbf{-997.2929}  & \textbf{-818.1903}  & \textbf{-750.7602}  &  \textbf{1492.2827}  & \textbf{388.5169}  \\[5pt]
            \multicolumn{3}{l}{\textsc{Multimodal functions}} \\[2pt]
            f6 & -900   & {-895.3502}  & {-821.1045}  &  {-822.4812} &  {-671.7844}  &  {52.6209}
               & -894.6910  & -834.4391  & -829.8255  &  -724.5643  & 47.7397  \\
            f7 & -800   & \textbf{-782.9525}  & \textbf{-683.7158}  & \textbf{-682.9493}  & \textbf{-467.0235}  & \textbf{56.9393}
               & -782.9525  & -694.4117  & -688.6216  & -467.0235   &  58.9648 \\
            f8 & -700   & -679.5066 & -679.1449  &  -679.1533  &  -678.8807  &  0.1561
               & \textbf{-679.5066}  &  \textbf{-679.1685} &  \textbf{-679.1596} &  \textbf{-678.9175}  &  \textbf{0.1489} \\
            f9 & -600   &  -595.5807 & -591.1572  & -591.6152  & -586.6314   & 2.1540
               & -595.8833  &  -591.6480 & -591.9006  &  -586.6852  & 2.2068  \\
            f10 & -500   &  -490.7566 & -456.7438  & -409.3499  &  275.8620  & 146.2619
                & {-495.2486}  &  {-447.6569} &  {-417.5570} &  {272.8506}  &  {119.5523} \\
            f11 & -400   & -381.8322  &  -350.7278 &  -350.8433  & -308.3906  & 16.9659
                &  -385.8263 &  -352.6462 & -349.7645  &  -306.3767  &  19.8457 \\
            f12 & -300   & -268.4056  & -236.9123  & -234.3349  &  -173.2333  &  21.4976
                & -267.5442  & -238.1376  &  -233.7328 &  -180.4475  &  24.0218 \\
            f13 & -200   & -164.4919  &  -135.5697 & -133.2558  &  -83.0992 & 16.5938  
                & -157.3840  & -133.8360  &  -130.9067  &  -101.3905 & 14.3010  \\
            f14 & -100   &  {426.6840}  &  {1445.3719} & {1372.1666}  &  {1919.8069}  & {371.0698} 
                &  529.2313 & 1437.8050  & 1408.9207  &  2354.3925  &  372.3278 \\
            f15 & 100   &   1220.1451 & 1947.0528  & 1975.5844  &  2562.6328  & 389.5177 
                & 1003.1814  & 1977.0000  &  1942.4554  & 2612.5980 & 420.3723 \\
            f16 & 200   &  200.9639 &  202.4348  &  202.4266 &  203.6991  &  0.7239 
                & 201.4183  &  202.5016 &  202.7041 &  204.6743  &  0.8043 \\
            f17 & 300   &  300.1525 & 301.0488  &  302.5322 &  344.4286  & 6.2663
                & 300.1525  &  301.1105 & 302.6090  &  344.4286  & 6.2806  \\
            f18 & 400   &  400.1525 &  401.0965 &  401.8234 & 406.8142   & 1.7579
                & {400.1525} & {401.0380} &  {401.8282} &  {409.3670} &  {1.9511}\\
            f19 & 500   & 503.2288  & 506.1112  & 507.2618  &  546.8400  &  6.1033
                & 503.2288  &  506.1112 & 506.5292  &   514.7202  &  2.2082 \\
            f20 & 600   & 603.7083  &  604.3811 & 604.3909  &  605.0000   &  0.3448
                &  603.7083  &  604.3823  &  604.3902  &   605.0000  & 0.3476  \\[5pt]
            \multicolumn{3}{l}{\textsc{Composition functions}} \\[2pt]
            f21 & 700   & 1023.3749  &  1513.9602 &  1479.0269 &  1552.7040  &  115.6359 &  1023.3749 & 1513.9602  &  1480.1213 &  1552.7040  & 115.4776  \\
            f22 & 800   &  1548.1995  &  2539.2318 & 2522.6101  &  3500.2226  & 530.8839  & \textbf{1438.2481}  & \textbf{2427.6843}  &   \textbf{2470.5047} &  \textbf{3502.0232} & \textbf{554.2658}  \\
            f23 & 900   & 2034.5278  & 2775.3844  &  2807.9612 &  3892.0847  & 378.4048  &  1969.5557 &   2863.4119 &  2857.0894  &  3605.0865  &  375.8073 \\
            f24 & 1000   &  1138.1878 & 1226.8297  & 1257.8455  &  1218.9173  & 25.6467  &  1118.7614 &   1222.2297 &  1202.6021 &  1235.4652  & 37.9750  \\
            f25 & 1100   &  1241.8700 &  1345.1502 & 1339.3068  &   1366.6899  &  25.4506  &  1240.1102 & 1346.0017  &  1337.4075 &  1365.8927  &  29.4815  \\
            f26 & 1200   &   1336.3902 &  1400.5789 &  1408.2959  & 1533.1272   &  49.3885 & 1331.5768  & 1400.6062  & 1407.4818  &  1529.5341  &   48.9864 \\
            f27 & 1300   &   1623.8144 & 1642.8999  &  1643.3209 &  1665.1437  &   7.4070 &  1623.8144 &  1641.5731 & 1642.7850  &  1665.1437  &  7.0987 \\
            f28 & 1400   & \textbf{1513.9127} &  \textbf{2153.4132} & \textbf{2071.3629} &  \textbf{2266.3627}  & \textbf{215.9328}
		   & 1513.9127  &  2149.7227 & 2071.4109  &  2266.3627   & 215.8366  \\\midrule
		   \multicolumn{7}{c|}{\textbf{SPSO2011}}   &  \multicolumn{5}{c}{\textbf{Bayesian Optimization}}\\\midrule
            \multicolumn{3}{l}{\textsc{Unimodal functions}} \\[2pt]
            f1 & -1400 & -1005.3148 & -103.485 & -7.8876 & 1695.6855 & 580.2353
                & 5216.1430 & 1.5510e+04 & 1.5117e+04 & 2.4693e+04 & 4857.8925 \\ 
            {f2} & -1300   & 2.3646e+07 & 1.3353e+08 & 1.4431e+08 & 2.6640e+08 & 6.6492e+07
               & 5.6854e+07 & 2.1022e+08 & 2.3433e+08 & 5.5868e+08 & 1.1446e+08 \\ 
            {f3} & -1200   &  1.2138e+10 & 8.9256e+10 & 1.06388e+11 & 2.7705e+11 & 5.7512e+10
               &  1.0004e+10 & 8.7289e+12 & 5.2365e+15 & 1.8529e+17 & 2.5868e+16 \\ 
            f4 & -1100   & 1.6502e+05 & 3.4671e+05 & 3.9683e+05 & 8.9498e+05 & 1.8831e+05
               & 4.6082e+04 & 9.1290e+05 & 4.1055e+06 & 2.9822e+07 & 6.2066e+06 \\ 
            f5 & -1000   & -831.6348 & -322.0166 & -308.399 & 505.4016 & 325.4384
               & 3434.6118 & 1.6727e+04 & 1.9594e+04 & 7.0332e+04 & 1.1099e+04  \\[5pt]
            \multicolumn{3}{l}{\textsc{Multimodal functions}} \\[2pt]
            f6 & -900   & -844.2954 & -770.1236 & -764.7432 & -674.2552 & 38.683
               &-572.7398 & 875.9278 & 1041.0426 & 4309.8598 & 937.4901 \\
            f7 & -800   & -701.8775 & -633.7445 & -632.2272 & -492.1828 & 43.0969
               & -605.0234 & 4392.9283 & 4.8434e+04 & 6.6221e+05 & 1.0378e+05 \\
            f8 & -700   & -679.4141 & -679.2812 & -679.2859 & -679.0626 & 0.1107
               & -679.3266 & -678.9201 & -678.9299 & -678.6414 & 0.1542 \\
            f9 & -600   &  -591.9697 & -589.397 & -589.4528 & -587.2924 & 0.9338
               & -590.7963 & -585.8143 & -585.8881 & -583.5415 & 1.353 \\
            f10 & -500   & -461.6521 & -274.8178 & -264.2497 & -0.613 & 101.7582
                & 176.7729 & 1733.1143 & 1824.9214 & 3644.3367 & 741.2554 \\
            f11 & -400   & -355.8199 & -308.9896 & -309.2968 & -271.9039 & 16.7635
                &  -253.1191 & -141.5103 & -137.2415 & 28.2658 & 59.2632 \\
            f12 & -300   & -248.9015 & -216.6082 & -215.6252 & -184.8619 & 14.6543
                & -146.2610 & -18.1314 & -18.5652 & 148.3443 & 66.9370 \\
            f13 & -200   &  -142.5769 & -110.8844 & -110.5187 & -71.8861 & 14.7131
                & -40.8029 & 73.0554 & 76.9612 & 263.1982 & 67.6236 \\
            f14 & -100   &  1580.9685 & 1957.0233 & 1957.8533 & 2308.8529 & 184.3534
                &  1763.2177 & 2647.0464 & 2657.2596 & 3247.7634 & 299.3089 \\
            f15 & 100   &  1509.0158 & 2251.4663 & 2220.2589 & 2576.8 & 216.8827
                & 1905.9186 & 2825.4242 & 2807.8911 & 3446.124 & 287.8989 \\
            f16 & 200   &  201.6575 & 202.6773 & 202.6513 & 203.8566 & 0.5514
                &  202.458 & 205.0509 & 205.0068 & 207.8333 & 1.3296 \\
            f17 & 300   &  374.5278 & 413.6145 & 416.1109 & 473.2804 & 20.4491
                & 564.6457 & 976.4061 & 960.0182 & 1343.7203 & 193.2775 \\
            f18 & 400   & 482.4333 & 518.1552 & 521.4023 & 565.5428 & 19.127
                & 664.6457 & 1076.4061 & 1060.6695 & 1443.7203 & 194.3157 \\
            f19 & 500   & 507.1073 & 544.8849 & 628.3692 & 2936.0275 & 341.7338
                & 3798.6579 & 1.5309e+05 & 2.7050e+05 & 1.2867e+06 & 2.8033e+05 \\
            f20 & 600   & 603.9779 & 604.2864 & 604.3053 & 604.6311 & 0.167
                & 604.8021 & 605.0000 & 604.9911 & 605.0000 & 0.0334 \\[5pt]
            \multicolumn{3}{l}{\textsc{Composition functions}} \\[2pt]
            f21 & 700   & 1117.9655 & 1172.2566 & 1176.7712 & 1292.8498 & 34.227
                & 2234.1194 & 2815.4961 & 2782.896 & 3827.9943 & 344.2327 \\
            f22 & 800   & 2515.5492 & 3119.0027 & 3090.0824 & 3447.205 & 197.5455
                &  2832.1769 & 3780.0713 & 3740.7229 & 4312.6797 & 293.4730 \\
            f23 & 900   & 2869.2363 & 3378.0357 & 3368.553 & 3697.4699 & 196.2726
                & 3087.3422 & 3818.1579 & 3771.9333 & 4278.5656 & 280.3538 \\
            f24 & 1000   & 1225.5428 & 1232.4759 & 1231.8447 & 1237.7855 & 3.3618
                & 1238.0768 & 1266.8067 & 1275.7245 & 1420.2028 & 32.5109 \\
            f25 & 1100   & 1323.2743 & 1330.8522 & 1330.7533 & 1338.3726 & 2.6161
                & 1355.4382 & 1372.0921 & 1375.891 & 1400.1419 & 12.1785 \\
            f26 & 1200   & 1364.9184 & 1395.6858 & 1398.6196 & 1521.2914 & 24.899
                & 1407.4966 & 1488.573 & 1493.4568 & 1565.5591 & 46.6492 \\
            f27 & 1300   & 1777.1847 & 1928.2138 & 1922.8312 & 2061.7246 & 69.009
                & 1647.6561 & 1696.2735 & 1711.6029 & 1967.458 & 51.9861 \\
            f28 & 1400   &  1880.9056 & 2359.1062 & 2346.6852 & 2469.9099 & 90.1052
                & 2798.8736 & 3475.094 & 3513.2157 & 4967.1594 & 405.5373  \\
            \bottomrule
            \end{tabular}}
            \caption{Performance of PSO with a GP-based heuristic on CEC2013 benchmark experiments.  Bold: Variant with lowest min value.}     
            \label{tab:computationalresultsC1C2SPSO2011BO}
\end{table}

\FloatBarrier
\subsection{Convergence plots}

\begin{figure}[H]
    \centering
    \thisfloatpagestyle{empty}
    \vspace{-4cm}
	\includegraphics[width=0.7\textwidth]{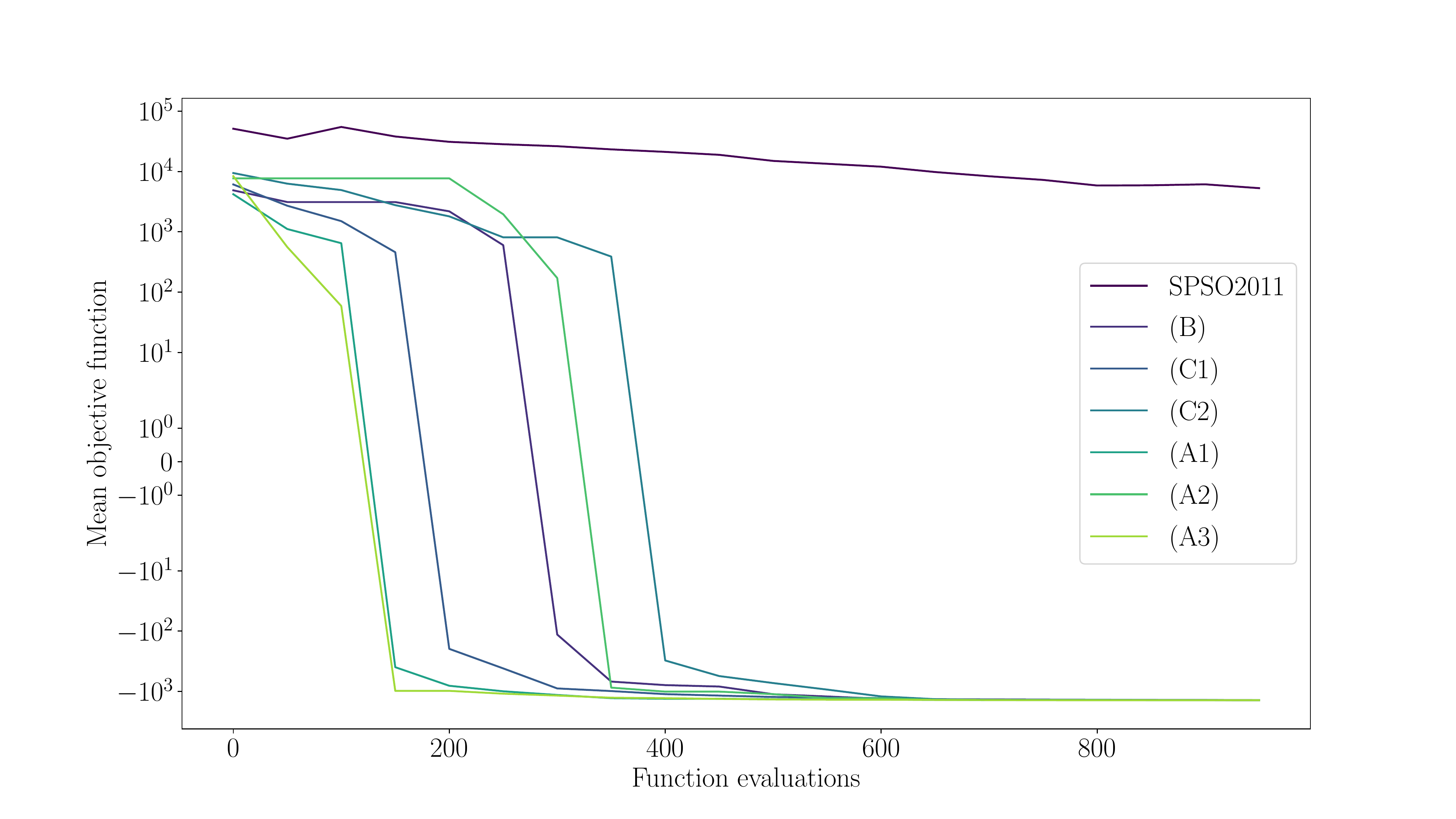}
	\includegraphics[width=0.7\textwidth]{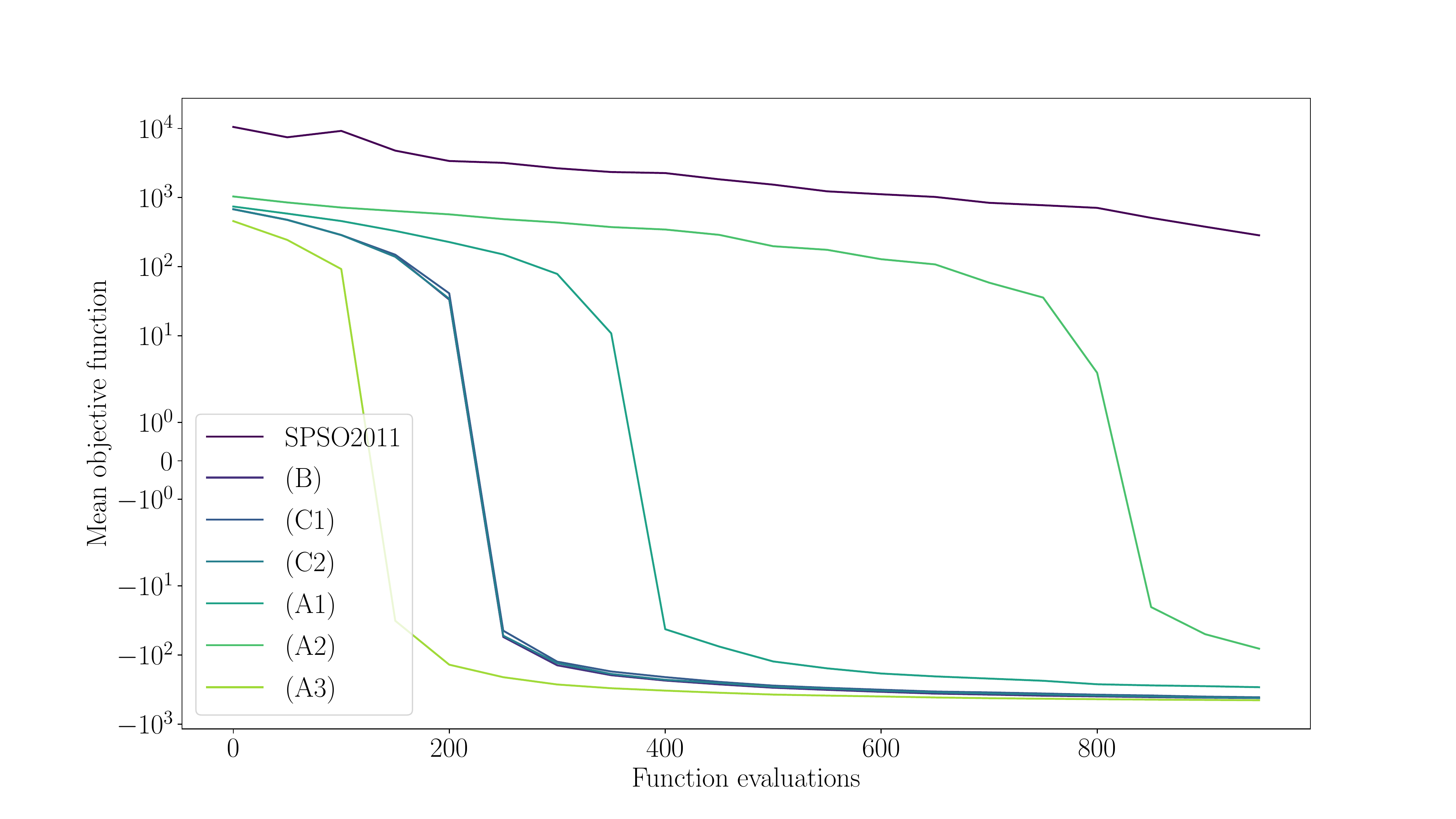}
	\includegraphics[width=0.7\textwidth]{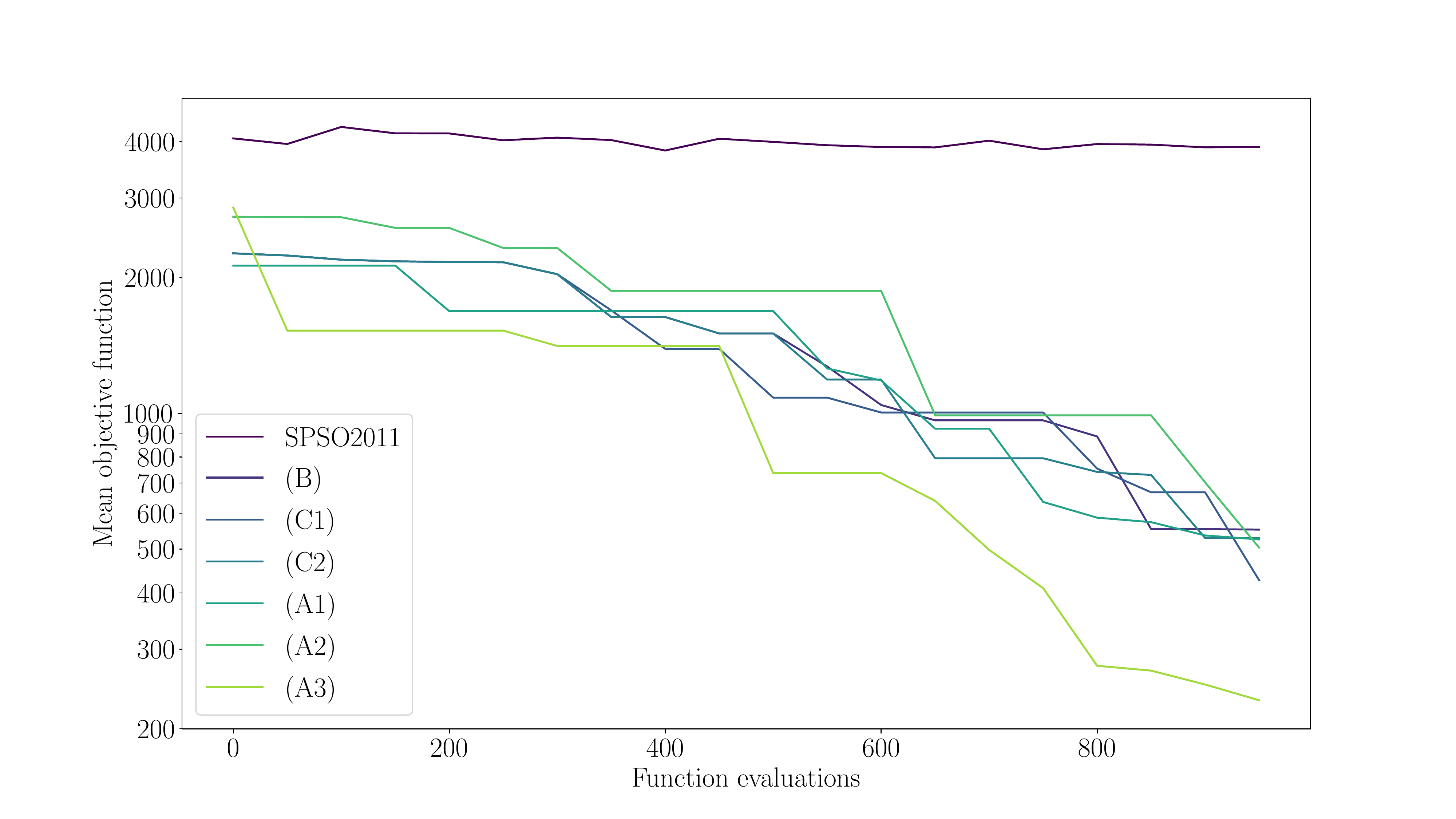}
	\includegraphics[width=0.7\textwidth]{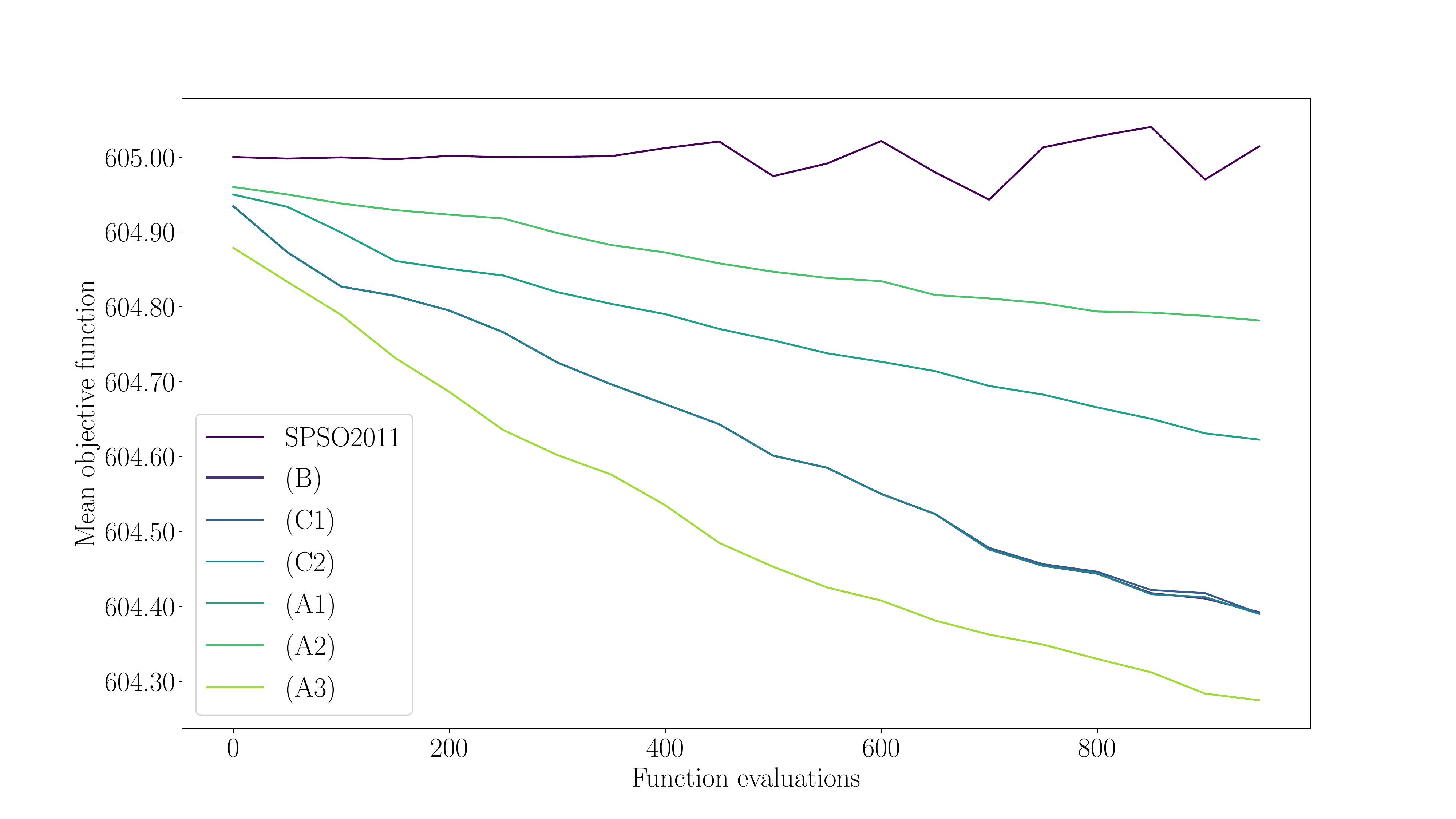}  
	\caption{Convergence plots for selected CEC2013 unimodal, multimodal and composition benchmark experiments (from top to bottom: f1: \textit{sphere function}; f10: \textit{rotated Griewank's function}; f14: \textit{Schwefel's function}; f20: \textit{expanded Schaffer’s F6 function}) for a limited number of function evaluations. Recall that above evaluation tables favored SPSO2011 over Bayesian Optimization and, hence, we exclusively focus on this benchmark.}
	\label{fig:convergence}
\end{figure}

\Cref{fig:convergence} depicts the convergence trajectory for selected CEC2013 benchmark experiments. Thereby, we provide further insights into the convergence behavior for unimodal, multimodal, and composition functions. We see that versions (A3), (B), and (C) improve the convergence and, in particular, outperform SPSO2011 (\eg, f10: rotated Griewank's function; f14: Schwefel's function). For all experiments, we yield improved candidate solutions in the initial search phase. Such behavior improves the effectiveness of the swarm search but, on top of that, is especially helpful in applications where function evaluations are costly (or even subject to budget limitations).  

\subsection{Runtime comparison}

\begin{figure}[h]
    \thisfloatpagestyle{empty}
    \centering
    \vspace{-4cm}    
	\includegraphics[width=0.7\textwidth]{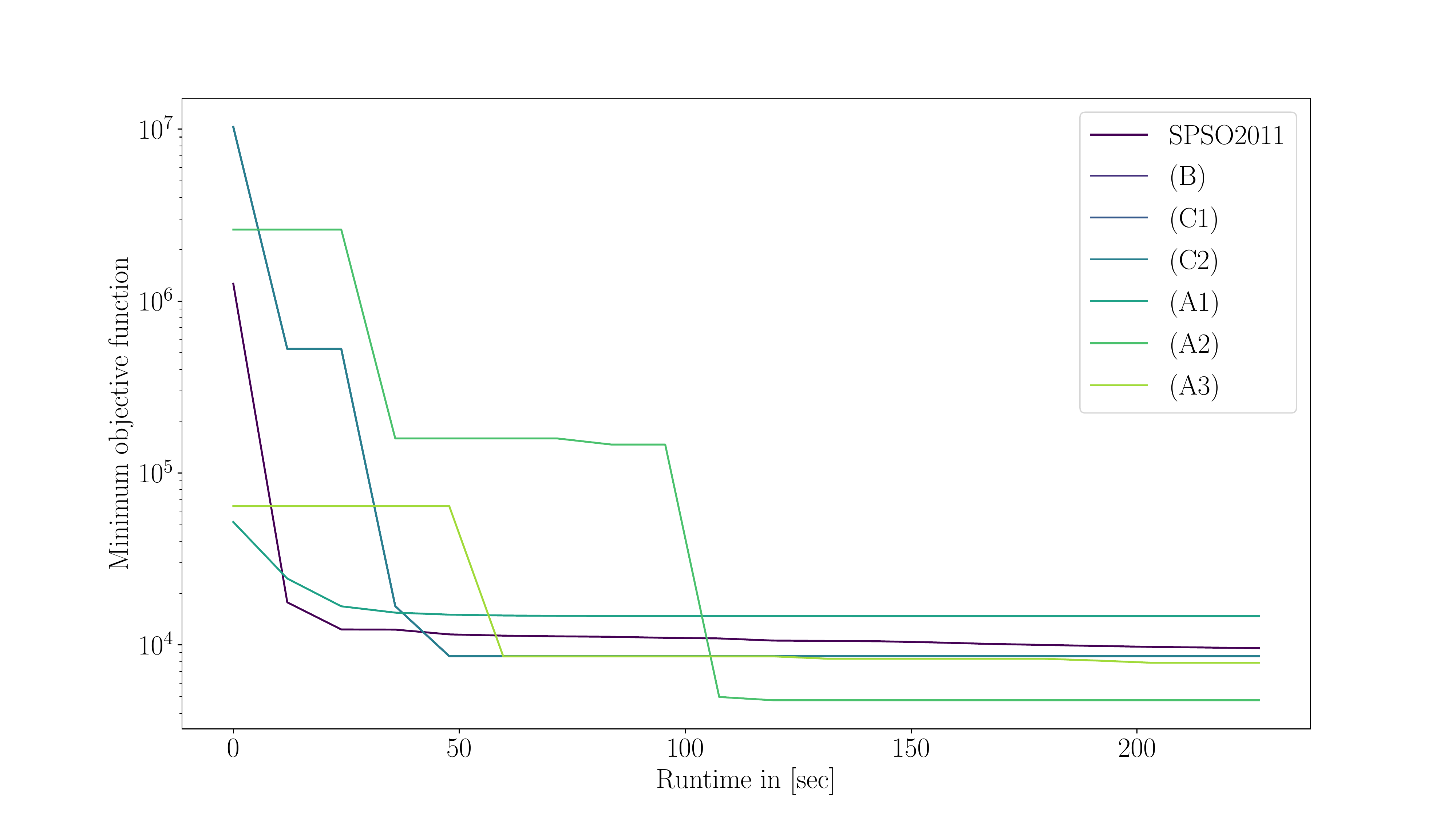}
	\includegraphics[width=0.7\textwidth]{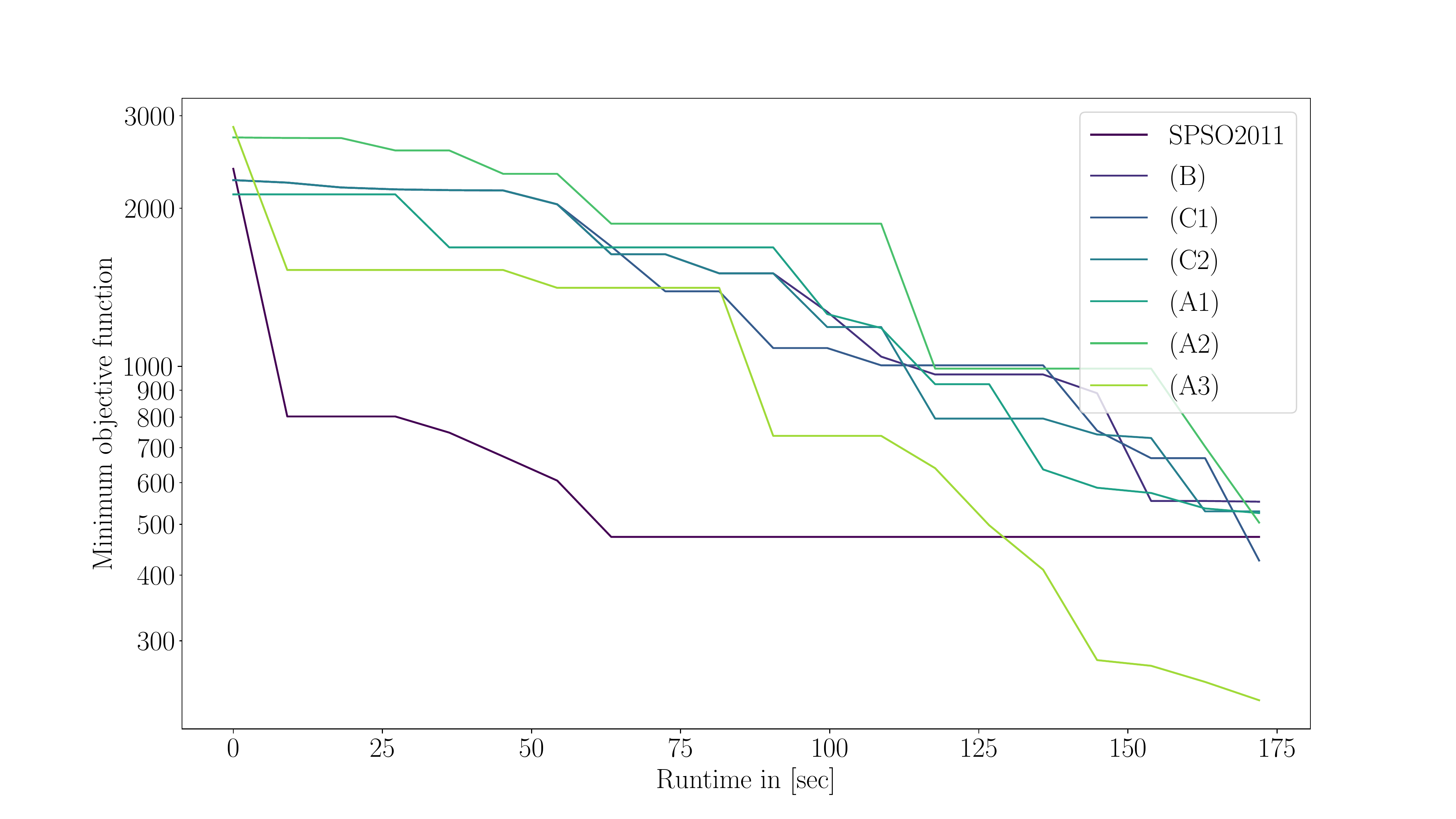}  
	\includegraphics[width=0.7\textwidth]{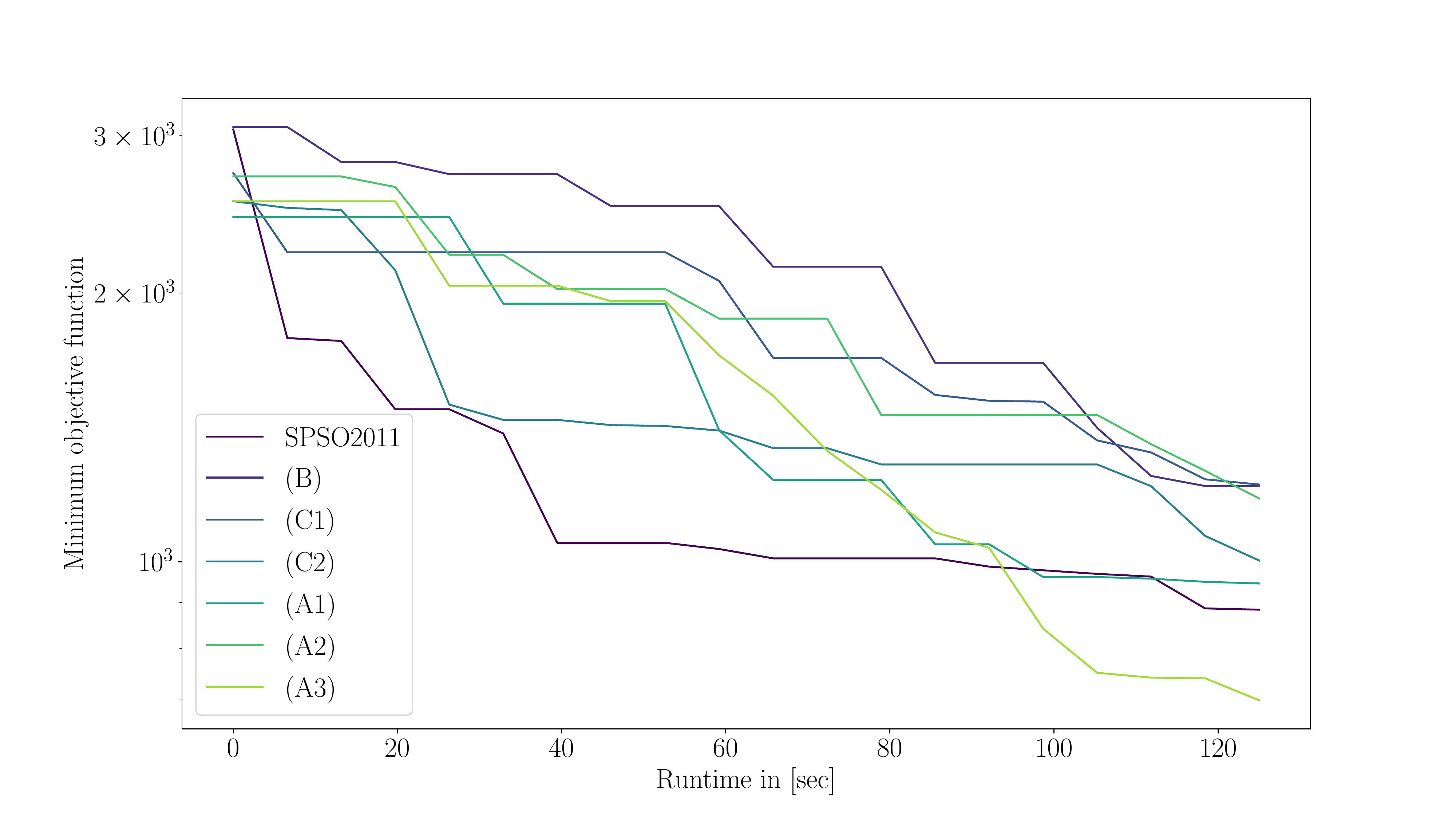}
	\includegraphics[width=0.7\textwidth]{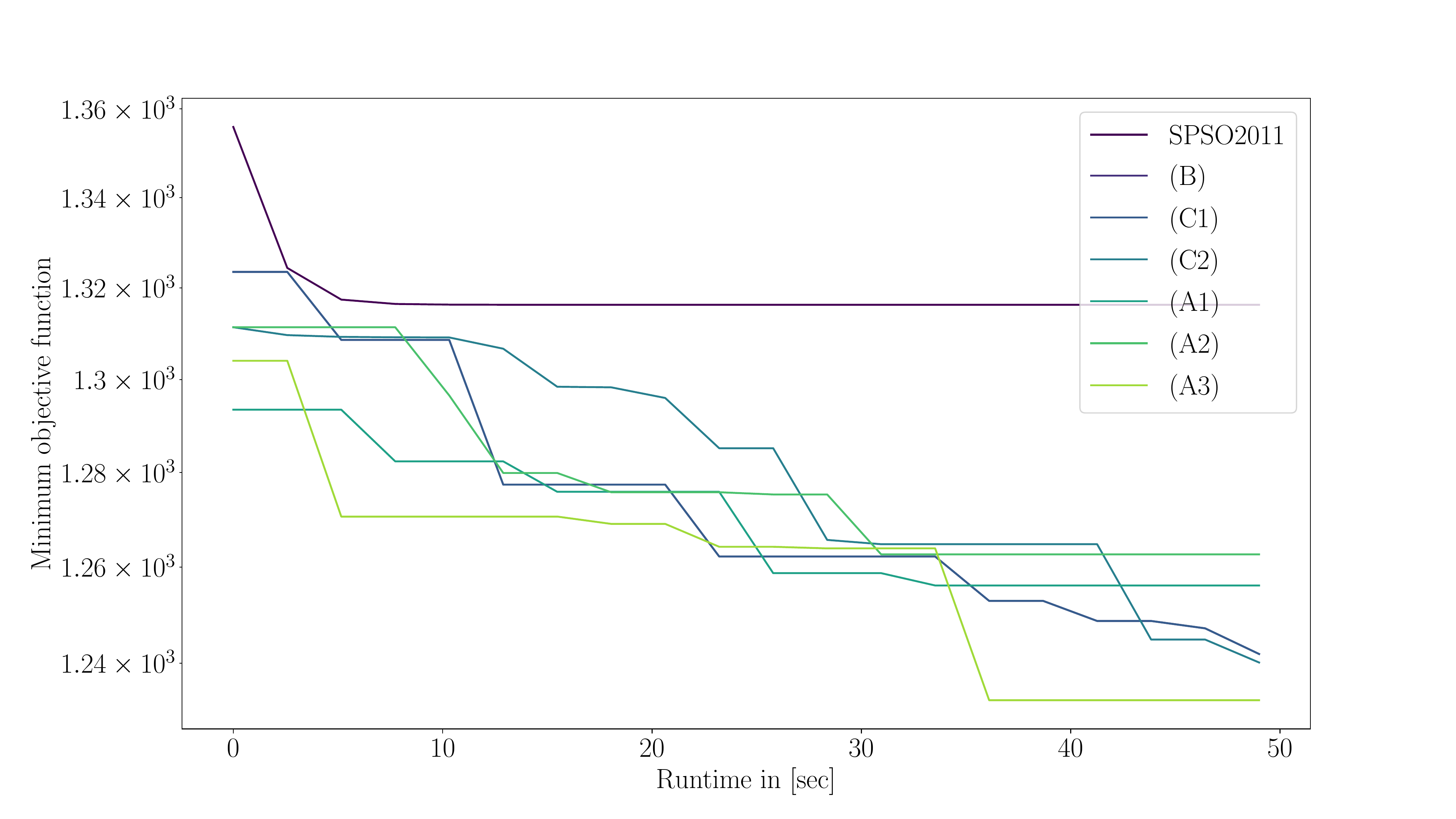}  
	\caption{Runtime comparison for selected CEC2013 experiments from unimodal, multimodal, and composition benchmarks (from top to bottom: f4: \textit{rotated discus function}; f14: \textit{Schwefel's function}; f15: \textit{rotated Schwefel's function}; f25: \textit{composition function 5 ($n= 3$, rotated)}). Experiments are performed on an Intel Core i5 with 3.1 GHz and 16\,GB memory. Bayesian optimization is omitted for reasons of space as it is already shown to be inferior in the above numerical experiments.}
	\label{fig:runtime}
\end{figure}

\Cref{fig:runtime} compares the runtime of SPSO2011 against our algorithms. On the one hand, our algorithms require additional computational time for estimating the GP, while, on the other hand, the GP helps in accelerating the search process. The following experiments are based on the actual runtime for evaluation $f$, which, in our experiments, is computationally cheap. However, we remind the reader that many applications of PSO in practice involve functions $f$ that are computationally costly and, therefore, the relative overhead due to the GP is likely to be lower in practice. 

As above, we provide results for unimodal, multimodal, and composition functions. In comparison to SPSO2011, variants (A3), (A2), (B), and (C) yield better solutions. The benefit of our algorithm over SPSO2011 becomes especially pronounced for objective functions of high complexity (\eg f14: Schwefel's function; f25: composition function 5 ($n= 3$, rotated)) where a directed search can drive the swarm towards better solution. This points towards the overall effectiveness of the directed search in our algorithm. 

\FloatBarrier
\subsection{Comparison against surrogate-assisted evolutionary algorithms}

We further compare our PSO variants against state-of-the-art surrogate-assisted evolutionary algorithms. Specifically, we compare against five popular variants: (1)~a Gaussian-process based genetic search \citep[GPEME;][]{liu2013gaussian}; (2)~an ensemble-based algorithm with reweighting according to surrogate errors \citep[WTA1;][]{goel2007ensemble}; (3)~a surrogate-assisted memetic algorithm that with surrogate-based local searches \citep[GS-SOMA;][]{lim2009generalizing}; (4)~a surrogate-assisted by local Gaussian random field metamodels \citep[MAES-ExI;][]{emmerich2006single} and (5)~a PSO extended by committee-based active learning \citep[CAL-SAPSO;][]{wang2017committee}. The evaluation setting is identical to prior literature \citet{wang2017committee}. This is done to allow for a fair comparison against existing surrogate-assisted evolution algorithms. Specifically, the evaluation is based on four popular multi-modal benchmark functions functions as in \citet{wang2017committee}: Ackley function, Griewank function, Rastrigin function, and Rosenbrock function. The performance of the algorithm is evaluated after $11 \cdot D$ function evaluations analogous to \citet{wang2017committee}. For reasons of space, we present the results for ten dimensional ($D=10$) objective functions. We report the best candidate solution values (mean) across 20 independent runs.

The results are in \Cref{tab:SAEAs}. We observe that all considered surrogate-assisted evolutionary algorithms are outperformed by variant~(A3) for two out of four functions. For instance, for the Ackley function, variant~(A3) registers a better result than the state-of-the-art benchmarks \citep[\ie, MEAS-ExI;][]{emmerich2006single} with an improvement of over 70\,\%. The Ackley function has a very narrow basin in the area of the global optimum, which we evaluate in the domain $\mathcal{D}=[-5, 5]^D$. On this function, non-ensemble-based techniques as, for instance, GPENE and MAES-ExI outperform ensemble-based methods (\eg, WTA1, GS-SOMA, CAL-SAPSO). Similarly, non-ensemble-based techniques perform better than ensemble-based methods on the multi-modal Rastrigin function for the domain $\mathcal{D}=[-5, 5]^D$ \citep[][]{wang2017committee}. In contrast to the surface of the Ackley function, the Rastrigin function includes a large set of local minima around the global optimum. For both functions, we observe that variant~(A3) outperforms existing non-ensemble-based methods (\ie GPMENE and MAES-ExI) and therefore achieves the overall best performance. Thus, the evaluation suggests that our proposed variant~(A3) is superior in the optimization performance of non-ensemble-based methods on objective functions where non-ensemble-based methods perform particularly strong.

We further benchmark our proposed variants on the Griewank function for $\mathcal{D}=[-600, 600]^D$ and on the Rosenbrock function for $\mathcal{D}=[-5, 5]^D$. While the Rosenbrock function includes a narrow valley leading to global optimum, the Griewank function comprises a large number of local minima around the global minimum. Our proposed variant~(A3) outperforms GPEME on the Griewank benchmark. The comparison against GPEME is particularly interesting, as GPEME is based upon a combination of Gaussian process and genetic algorithm. Here we find that variant~(A3) is superior in three out of four cases. This can directly attributed to the difference between both: GPEME uses the surrogate only for prescreening candidate solutions, whereas we actually integrate the surrogate to guide the swarm search. Overall, this establishes the strong performance of the proposed algorithms as they widely preferred.

\begin{table}[]
    \centering
    \thisfloatpagestyle{plain}
    \scriptsize
    \renewcommand{\arraystretch}{1}
    \centerline{
    \begin{tabular}{>{}l>{}c>{}c>{}c>{}c}
            \toprule
            \textbf{Algorithm} & \multicolumn{4}{l}{\textbf{Benchmarks}}\\
            & \textbf{Griewank} \citep[][]{griewank1981generalized} & \textbf{Rosenbrock} \citep{rosenbrock1960automatic} & \textbf{Rastrigin} \citep[][]{rastrigin1974systems} & \textbf{Ackley} \citep[][]{ackley2012connectionist} \\
            \toprule
            \multicolumn{3}{l}{\textsc{Proposed variants from this paper}} \\[2pt]
            Variant (A1) &  6.40e+00($\pm$3.19e+00) &  1.05e+03($\pm$1.18e+03) &  8.04e+01($\pm$2.14e+01) &  3.42e+00($\pm$2.14e+00) \\
            Variant (A2) &  1.68e+01($\pm$2.21e+01) &  4.56e+03($\pm$1.30e+04) &  9.71e+01($\pm$2.74e+01) &  5.29e+00($\pm$2.48e+00) \\
            \textbf{Variant (A3)} &  4.53e+00($\pm$2.15e+00) &  1.12e+03($\pm$1.81e+03) &  \textbf{6.85e+01($\pm$1.87e+01)} &  \textbf{2.05e+00($\pm$6.33e--01)} \\
            Variant (B)  &  3.60e+01($\pm$2.14e+01) &  1.27e+03($\pm$1.00e+03) &  7.82e+01($\pm$2.02e+01) &  4.96e+00($\pm$1.04e+00) \\
            Variant (C1) &  3.43e+01($\pm$2.14e+01) &  1.31e+03($\pm$1.11e+03) &  7.81e+01($\pm$1.99e+01) &  5.04e+00($\pm$9.92e--01) \\
            Variant (C2) &  3.05e+01($\pm$1.85e+01) &  1.39e+03($\pm$1.49e+03) &  7.70e+01($\pm$2.10e+01) &  4.94e+00($\pm$1.06e+00) \\[5pt]
            \multicolumn{3}{l}{\textsc{Benchmark algorithms}} \\[2pt]
            SPSO2011 & 5.50e+01($\pm$1.72e+01) & 7.90e+03($\pm$5.57e+03) & 9.95e+01$\pm$1.60e+01) & 1.65e+01($\pm$1.65e+00) \\
            BO & 4.43e+01($\pm$7.48e+01) & 3.44e+09($\pm$1.90e+09) & 1.38e+04$\pm$3.01e+03) & 2.11e+01($\pm$1.35e--01) \\[5pt]
            \multicolumn{3}{l}{\textsc{State-of-the-art surrogate-assisted evolutionary algorithms (SAEAs)}} \\[2pt]
            CAL-SAPSO 
            & 1.12e+00 ($\pm$ 1.21e--01) & \textbf{1.77e+00($\pm$3.80e--01)} & 8.88e+01($\pm$2.26e+01) &  2.01e+01($\pm$2.44e--01) \\
            WTA1 
            & \textbf{1.07e+00 ($\pm$1.04e--02)} & 1.18e+01($\pm$2.13e--01) & 9.58e+01($\pm$3.20e+00) &  1.90e+01($\pm$1.23e+00)\\
            GS-SOMA 
            & 1.08e+00 ($\pm$1.78e--01) & 4.77e+00($\pm$1.14e+00) & 1.05e+02($\pm$1.52e+01) &  1.84e+01($\pm$1.73e+00) \\
            GPEME 
            & 2.72e+01 ($\pm$1.13e+01) & 2.07e+01($\pm$7.44e+00) & 7.15e+01($\pm$1.27e+01) &  1.38e+01($\pm$2.50e+00) \\
            MEAS-ExI 
            & 1.20e+01 ($\pm$5.19e+00) & 1.69e+01($\pm$4.63e+00) & 8.75e+01($\pm$1.29e+01) &  7.49e+00($\pm$3.77e+00) \\\bottomrule
    \end{tabular}}
    \caption{Comparison of proposed variants in this paper against surrogate-assisted evolutionary algorithms. The setting is chosen as in other surrogate-assisted evolutionary algorithms for $D=10$ \citep{wang2017committee}. Here the corresponding objective function was minimized. Results were obtained across 20 runs and, across all runs, report is the mean over the best candidate solution were averaged. Reported is then the mean ($\pm$SD).}
    \label{tab:SAEAs}
\end{table}

\FloatBarrier
\section{Discussion}
\label{sec:discussion}


As shown above, the proposed algorithms improves the efficiency over standard PSO, that is, with the same number of iterations, a better solution is obtained. This is especially crucial in practical settings where function evaluations are economically or computationally costly \citep[cf.][for a discussion]{Rios.2013}. Fr potential users of our algorithms, there are direct implications. In a cost analysis, we showed that the benefits of our algorithm relate to the cost (or runtime) of making function evaluations. If function evaluations are (computationally) cheap, a standard PSO appears beneficial, while, otherwise, our Gaussian-process-based PSO has benefits. Similarly, the benefits of our algorithms over standard PSO diminish if many function evaluations can be made, yet, when this is limited as in the above experiments, our algorithms are favorable.


Our work developed variations to PSO by using the information gained about the underlying function to model this function itself as a stochastic process. In other words, this could also be seen as an extension of concepts from Bayesian optimization to PSO. By drawing upon Gaussian processes, we allow for a wide family of possible function structures and benefit from recent advances in Bayesian optimization that currently serves as a popular algorithm in high-expense black-box function optimization. Along with variant~(A3), variant~(C) with a combination of PSO and GP-based exploitation showed promising results in the numerical experiments. Besides that, this variant has interest theoretical properties, as it yields the same performance guarantees as sequential Bayesian optimization (subject to a fixed factor). As opposed to pure Bayesian optimization, the PSO search remains agnostic to the underlying function. It might therefore serve as an initial exploratory scheme until the function can be adequately modeled. Here we point to \citet{Buche.2005}, who use an evolutionary algorithm followed by Bayesian optimization in similar fashion.

\section{Conclusion}
\label{sec:conclusion}


This work proposed an innovative direction for improving stochastic optimization methods by guiding the search process with a stochastic function approximation. Specifically, we combined particle swarm optimization with Gaussian-process-based function approximation between the walkers' positions in order to better improve particle movements with regard to exploration and exploitation. This allows us to profit from the stochastic convergence of the swarm, while also accelerating convergence and reducing risks of finding local instead of global optima. To this end, three different approaches were developed: (A)~using the stochastic surrogate model to locate an approximate global optimum, and include this in the original PSO search direction; (B)~using the approximate global optimum to leapfrog the worst particle for fast convergence; (C)~a hybridization of PSO and Bayesian optimization for increased exploitation. As a result, these modifications can improve the stochastic search process over the original PSO by obtaining a more effective swarm behavior with respect exploration and exploitation. Finally, our computational experiments demonstrated that enhancing PSO by leveraging a Gaussian-process-bases surrogate model as a further source of information improves its performance. 


The above method opens avenues for future research. Specifically, one could investigate the performance when replacing the GP approximation with alternative models, such as Bayesian neural networks, random forests, or other models that allow for a stochastic interpretation of functions. Here one could also decompose the approximation into a stochastic one for uncertainty estimates and a non-stochastic counterpart for the response surface, as this could empower more accurate approximations at lower computational costs. Our approach of a GP-guided stochastic optimization might also be helpful for other population-based algorithms that involve a combination of exploratory and exploitative behavior (\eg, evolutionary algorithms, cross-entropy methods and bees algorithms). Beyond that, we leave it to future research to adapt the proposed algorithm to parallel multiple swarm optimization and multi-objective optimization. 

\section*{Acknowledgment}

\footnotesize Cloud computing resources were provided by a Microsoft Azure for Research award. We thank the review team for valuable suggestions, in particular with regard to the experimental setup.

\setlength{\bibsep}{0.15\baselineskip}
{
\setlength{\bibsep}{3pt}
\bibliographystyle{model5-names-no-doi}
\bibliography{literature.bib}
}

\newpage
\appendix
\pagestyle{empty}

\begin{center}
\Large
\textbf{Online Supplements}
\end{center}
\normalsize

\section{CEC2013 benchmark functions}
\label{appendix:cec2013}
\setcounter{table}{0}

\begin{table}[htb]
    \thisfloatpagestyle{plain}
	\centering
	\scriptsize
	\renewcommand{\arraystretch}{0.618}
	\begin{tabular}{lSc}
		\toprule
		\textbf{Function} & \multicolumn{1}{c}{\textbf{Optimum}} & \textbf{Domain}\\
		\midrule
		\multicolumn{2}{l}{\textsc{Unimodal functions}} \\[5pt]
		f1 \quad Sphere function  & -1400 & $\intervalcc{-100}{100}^D$\\
		f2 \quad Rotated high conditioned elliptic function  & -1300 & $\intervalcc{-100}{100}^D$ \\
		f3 \quad Rotated bent cigar function & -1200 & $\intervalcc{-100}{100}^D$ \\
		f4 \quad Rotated discus function  & -1100 & $\intervalcc{-100}{100}^D$\\
		f5 \quad Different powers function  & -1000 & $\intervalcc{-100}{100}^D$\\[5pt]
		\multicolumn{2}{l}{\textsc{Multimodal functions}} \\[5pt]
		f6 \quad Rotated Rosenbrock’s function  & -900 & $\intervalcc{-100}{100}^D$ \\
		f7 \quad Rotated Schaffers F7 function  & -800 & $\intervalcc{-100}{100}^D$\\
		f8 \quad Rotated Ackley’s function  & -700 & $\intervalcc{-100}{100}^D$\\
		f9 \quad Rotated Weierstrass function & -600 & $\intervalcc{-100}{100}^D$\\
		f10 \quad Rotated Griewank’s function & -500 & $\intervalcc{-100}{100}^D$\\
		f11 \quad Rastrigin’s Function & -400 & $\intervalcc{-100}{100}^D$\\
		f12 \quad Rotated Rastrigin’s function & -300 & $\intervalcc{-100}{100}^D$\\
		f13 \quad Non-continuous rotated Rastrigin’s function & -200 & $\intervalcc{-100}{100}^D$\\
		f14 \quad Schwefel's function & -100 & $\intervalcc{-100}{100}^D$\\
		f15 \quad Rotated Schwefel's function & 100 & $\intervalcc{-100}{100}^D$\\
		f16 \quad Rotated Katsuura function & 200 & $\intervalcc{-100}{100}^{D}$\\
		f17 \quad Lunacek Bi-Rastrigin function & 300 & $\intervalcc{-100}{100}^{D}$ \\
		f18 \quad Rotated Lunacek Bi-Rastrigin function & 400 & $\intervalcc{-100}{100}^{D}$ \\
		f19 \quad Expanded Griewank’s plus Rosenbrock’s function & 500 & $\intervalcc{-100}{100}^{D}$\\
		f20 \quad Expanded Schaffer’s F6 Function & 600 & $\intervalcc{-100}{100}^{D}$\\[5pt]
		\multicolumn{2}{l}{\textsc{Composition functions}} \\[5pt]
		f21 \quad Composition function 1 ($n=5$, rotated)  & 700 & $\intervalcc{-100}{100}^{D}$\\
	    f22 \quad Composition function 2 ($n=3$, unrotated)  & 800 & $\intervalcc{-100}{100}^{D}$\\
		f23 \quad Composition function 3 ($n=3$, rotated)  & 900 & $\intervalcc{-100}{100}^{D}$\\
		f24 \quad Composition function 4 ($n=3$, rotated)  & 1000 & $\intervalcc{-100}{100}^{D}$\\
		f25 \quad Composition function 5 ($n=3$, rotated)  & 1100 & $\intervalcc{-100}{100}^{D}$\\
		f26 \quad Composition function 6 ($n=5$, rotated)  & 1200 & $\intervalcc{-100}{100}^{D}$\\
		f27 \quad Composition function 7 ($n=5$, rotated)  & 1300 & $\intervalcc{-100}{100}^{D}$\\
		f28 \quad Composition function 8 ($n=5$, rotated)  & 1400 & $\intervalcc{-100}{100}^{D}$\\
		\bottomrule
	\end{tabular}
	\caption{Evaluation functions from CEC2013 benchmark suite \citep{liang2013problem}.}
	\label{tab:functions}
\end{table}

\clearpage
\section{Statistical tests for performance comparisons}
\label{appendix:stattests}
\setcounter{table}{0}

We report $p$-values for the significance of the results presented in \Cref{tab:computationalresultsA1A2A3B,tab:computationalresultsC1C2SPSO2011BO}. To this end, we calculate $p$-values of a one-sided $t$-test on the means of favored variant~(A3) versus the remaining variants and the benchmark algorithms. 

\begin{table}[htb]
    \thisfloatpagestyle{plain}
	\centering
	\scriptsize
	\renewcommand{\arraystretch}{0.618}
	\sisetup{detect-all}
	\centerline{
	\begin{tabular}{lSSSSSSSS}
		\toprule
		\textbf{Function} & \textbf{(A3)} & \multicolumn{1}{c}{\textbf{(A1)}} & \textbf{(A2)} & \textbf{(B)}& \textbf{(C1)}& \textbf{(C2)}& \textbf{(SPSO2011)}& \textbf{(BO)}\\
		\midrule
		\multicolumn{5}{l}{\textsc{Unimodal functions}} \\[5pt]
		f1 & 1.0 & 0.0726 & 0.1423 & 0.0735 & 0.0867 & \bfseries 0.0049 & \bfseries 0.0000 & \bfseries 0.0000 \\
		2 & 1.0 & \bfseries 0.0020 & \bfseries 0.0000 & 0.2400 & 0.2118 & 0.2364 & \bfseries 0.0000 & \bfseries 0.0000\\
		f3 & 1.0 & 0.0552 & \bfseries 0.0346 & 0.1718 & 0.1718 & 0.1494 & \bfseries 0.0000 & 0.1103 \\
		f4 & 1.0 & \bfseries 0.0299 & 0.0674 & 0.0841 & 0.0860 & \bfseries 0.0430 & \bfseries 0.0261 & \bfseries 0.0014 \\
		f5 & 1.0 & \bfseries 0.0009 & \bfseries 0.0000 & \bfseries 0.0000 & \bfseries 0.0002 & \bfseries 0.0001 & \bfseries 0.0000 & \bfseries 0.0000 \\[15pt]
		\multicolumn{5}{l}{\textsc{Multimodal functions}} \\[5pt]
		f6 & 1.0 & \bfseries 0.0001 & \bfseries 0.0000 & \bfseries 0.0024 & \bfseries 0.0027 & \bfseries 0.0161 & \bfseries 0.0000 & \bfseries 0.0000 \\
		f7 & 1.0 & \bfseries 0.0030 & \bfseries 0.0001 & 0.4356 & 0.4356 & 0.3981 & \bfseries 0.0000 & \bfseries 0.0006 \\
		f8 & 1.0 & \bfseries 0.0001 & \bfseries 0.0000 & 0.0604 & 0.1232 & 0.1680 & \bfseries 0.0000 & \bfseries 0.0000 \\
		f9 & 1.0 & \bfseries 0.0000 & \bfseries 0.0000 & \bfseries 0.0142 & \bfseries 0.0000 & \bfseries 0.0000 & \bfseries 0.0000 & \bfseries 0.0000 \\
		f10 & 1.0 & \bfseries 0.0001 & \bfseries 0.0000 & \bfseries 0.0371 & \bfseries 0.0283 & \bfseries 0.0337 & \bfseries 0.0000 & \bfseries 0.0000 \\
		f11 & 1.0 & \bfseries 0.0000 & \bfseries 0.0000 & \bfseries 0.0173 & 0.0616 & \bfseries 0.0499 & \bfseries 0.0000 & \bfseries 0.0000 \\
		f12 & 1.0 & \bfseries 0.0000 & \bfseries 0.0000 & \bfseries 0.0025 & \bfseries 0.0002 & \bfseries 0.0005 & \bfseries 0.0000 & \bfseries 0.0000 \\
		f13 & 1.0 & \bfseries 0.0000 & \bfseries 0.0000 & \bfseries 0.0001 & \bfseries 0.0000 & \bfseries 0.0000 & \bfseries 0.0000 & \bfseries 0.0000 \\
		f14 & 1.0 & \bfseries 0.0001 & \bfseries 0.0000 & \bfseries 0.0007 & \bfseries 0.0028 & \bfseries 0.0004 & \bfseries 0.0000 & \bfseries 0.0000 \\
		f15 & 1.0 & \bfseries 0.0004 & \bfseries 0.0000 & \bfseries 0.0002 & \bfseries 0.0008 & \bfseries 0.0001 & \bfseries 0.0000 & \bfseries 0.0000 \\
		f16 & 1.0 & \bfseries 0.0002 & 0.2957 & 0.0511 & 0.2957 & 0.0692 & \bfseries 0.0000 & \bfseries 0.0000\\
		f17 & 1.0 & \bfseries 0.0149 & \bfseries 0.0029 & 0.1500 & 0.2873 & 0.2651 & \bfseries 0.0000 & \bfseries 0.0000\\
		f18 & 1.0 & \bfseries 0.0003 & \bfseries 0.0003 & 0.1294 & 0.2845 & 0.2875 & \bfseries 0.0000 & \bfseries 0.0000\\
		f19 & 1.0 & \bfseries 0.0433 & \bfseries 0.0007 & 0.2666 & 0.4853 & 0.2702 & \bfseries 0.0036 & \bfseries 0.0000 \\
		f20 & 1.0 & \bfseries 0.0000 & \bfseries 0.0000 & 0.0752 & 0.2118 & 0.0796 & \bfseries 0.0000 & \bfseries 0.0000\\[15pt]
		\multicolumn{5}{l}{\textsc{Composition functions}} \\[5pt]
		f21 & 1.0 & \bfseries 0.0000 & \bfseries 0.0000 & 0.4116 & 0.4141 & 0.3961 & \bfseries 0.0000 & \bfseries 0.0000\\
		f22 & 1.0 & \bfseries 0.0000 & \bfseries 0.0000 & \bfseries 0.0120 & \bfseries 0.0023 & \bfseries 0.0098 & \bfseries 0.0000 & \bfseries 0.0000\\
		f23 & 1.0 & \bfseries 0.0020 & \bfseries 0.0000 & \bfseries 0.0166 & \bfseries 0.03995 & \bfseries 0.0077 & \bfseries 0.0000 & \bfseries 0.0000\\
		f24 & 1.0 & \bfseries 0.0052 & \bfseries 0.0000 & 0.0731 & 0.0560 & 0.2303 & \bfseries 0.0000 & \bfseries 0.0000\\
		f25 & 1.0 & \bfseries 0.0398 & \bfseries 0.0000 & 0.1677 & 0.2042 & 0.3392 & \bfseries 0.0000 & \bfseries 0.0000 \\
		f26 & 1.0 & 0.0963 & \bfseries 0.0048 & 0.4890 & 0.4406 & 0.4095 & 0.2308 & \bfseries 0.0000 \\
		f27 & 1.0 & \bfseries 0.0000 & \bfseries 0.0000 & 0.1060 & 0.0994 & 0.1841 & \bfseries 0.0000 & \bfseries 0.0000\\
		f28 & 1.0 & \bfseries 0.0235 & \bfseries 0.0000 & 0.4566 & 0.2630 & 0.2626 & \bfseries 0.0000 & \bfseries 0.0000\\
		\bottomrule
	\end{tabular}}
	\caption{Computed $p$-values of one-sided $t$-tests checking if the means of variant~(A3) are significantly lower than the means of the remaining variants and benchmark algorithms for the CEC2013 functions. We highlight $p$-values below a significance level of 5\% in bold.}
	\label{tab:pvalues}
\end{table}

\newpage
\section{Bayesian optimization}
\label{appendix:bayesian_optimization}


Bayesian optimization \citep{Mockus.1975} aims at finding the global optimum of a (non-convex) function $f: \domain \to \R$. The underlying idea is a stochastic model of the black-box function $f$. For this reason, one sequentially chooses a candidate solution $x_n$ that promises the highest gain according to a so-called \emph{acquisition function} $R: \domain \to \R$, \ie $x_n = \argmax_{x \in \domain} R(x)$ depending on the current stochastic model of the function. After evaluating the function at this location, one updates the model according to Bayes' rule. This process is repeated until convergence. This idea goes back at least to \citet{Kushner.1964}.

The above approach requires a stochastic model of the function. A common choice is the family of Gaussian processes, due to their computational tractability as seen in \Cref{sec:gp}. This family has been used in several approaches to Bayesian optimization, such as \citet{Jones.1998,Snoek.2012} or \citet{MoralesEnciso.2015} with various choices of kernels. 

Mathematically speaking, at each time step $n$, one defines a probabilistic model $f \sim \mathcal{P}^{(n)}$, based on the past observations. In the case of Gaussian processes, this is simply the posterior Gaussian process conditional on the past observations. The acquisition function $R$ is then defined based on this distribution. Examples are the probability of improvement \citep{Kushner.1964}, the expected improvement $\expectation_{\mathcal{P}^{(n)}} f(x) \indicator[x>y]$ \citep{Jones.1998,Mockus.1975} and the upper/lower confidence bound \citep{Auer.2002,Srinivas.2012}. 

For the case of Bayesian optimization using Gaussian processes, convergence guarantees have been made for expected improvement when using a fixed covariance function under certain assumptions \citep{Bull.2011}.

\end{document}